\documentclass[final,5p,times,twocolumn]{elsarticle}

\usepackage{amssymb}
\usepackage{amsmath}

\usepackage{textcomp}
\usepackage{stfloats}
\usepackage{url}
\usepackage{verbatim}
\usepackage[utf8]{inputenc}
\usepackage{booktabs}
\usepackage{xcolor}
\usepackage{amsmath}
\usepackage{multirow}
\usepackage{graphicx}    
\usepackage{caption}     
\usepackage{subcaption}  
\usepackage{adjustbox}
\usepackage{xcolor}
\usepackage{longtable}
\usepackage{array}
\usepackage{multirow}

\usepackage{rotating}

\usepackage{hyperref}

\definecolor{bestcolor}{HTML}{BE830E}
\definecolor{ANUGold}{HTML}{BE830E}
\definecolor{ANUGoldTint}{HTML}{F5EDDE}
\definecolor{ANUGoldDark}{HTML}{9A6B0B}
\definecolor{ANUGoldLight}{HTML}{D4A656} 
\definecolor{ANUUnigrey}{HTML}{333333}
\definecolor{ANUBlack}{HTML}{000000}  
\definecolor{best}{HTML}{ED6F6E} 
\definecolor{second}{HTML}{5560AC} 

\journal{Pattern Recognition}

\begin{document}

\begin{frontmatter}



\title{Representation-Centric Survey of {Supervised} Skeletal Action Recognition \\and the New Benchmark}


\author[ANU]{Yang Liu\fnref{equal}}
\author[UAB]{Jiyao Yang\fnref{equal}}
\author[Data61]{Madhawa Perera}
\author[OPPO]{Pan Ji}
\author[POSTECH]{Dongwoo Kim}
\author[CMU]{Min Xu}
\author[UAB]{Tianyang Wang}
\author[UWA]{Saeed Anwar}
\author[Curtin]{Tom Gedeon}
\author[Griffith,Data61]{Lei Wang\corref{cor1}}
\ead{l.wang4@griffith.edu.au}
\author[ANU,Yale]{Zhenyue Qin\corref{cor1}}
\ead{zhenyue.qin@yale.edu}

\fntext[equal]{Yang Liu and Jiyao Yang contributed equally to this work.}
\cortext[cor1]{Corresponding authors.}

\affiliation[ANU]{organization={School of Computing, Australian National University},
            city={Canberra},
            state={ACT},
            country={Australia}}

\affiliation[UAB]{organization={University of Alabama at Birmingham},
            city={Birmingham},
            state={AL},
            country={USA}}

\affiliation[Data61]{organization={Data61/CSIRO},
            city={Canberra},
            state={ACT},
            country={Australia}}

\affiliation[OPPO]{organization={OPPO US Research Center},
            city={Palo Alto},
            state={CA},
            country={USA}}

\affiliation[POSTECH]{organization={POSTECH},
            city={Pohang},
            country={South Korea}}

\affiliation[CMU]{organization={Carnegie Mellon University},
            city={Pittsburgh},
            state={PA},
            country={USA}}

\affiliation[UWA]{organization={University of Western Australia},
            city={Perth},
            state={WA},
            country={Australia}}

\affiliation[Curtin]{organization={Curtin University},
            city={Perth},
            state={WA},
            country={Australia}}

\affiliation[Griffith]{organization={School of Engineering and Built Environment, Griffith University},
            city={Brisbane},
            state={QLD},
            country={Australia}}

\affiliation[Yale]{organization={School of Medicine, Yale University},
            city={New Haven},
            state={CT},
            country={USA}}
\begin{abstract}
3D {skeletal} action recognition has emerged as a powerful alternative to traditional RGB and depth-based approaches, offering robustness to environmental variations, computational efficiency, and enhanced privacy. Despite remarkable progress, current research remains fragmented across diverse input representations and lacks evaluation under scenarios that reflect real-world challenges.
This paper presents a representation-centric review of {supervised skeletal action recognition}, systematically categorizing state-of-the-art methods by their input feature types: joint coordinates, bone vectors, motion flows, and extended representations, and analyzing how these choices influence spatiotemporal modeling strategies. 
Building on the insights from this review, we introduce ANUBIS, a large-scale, challenging {dataset} designed to address critical gaps in existing benchmarks. ANUBIS incorporates multi-view recordings with back-view perspectives, complex multi-person interactions, fine-grained and violent actions, and contemporary social behaviors.
We benchmark a diverse set of state-of-the-art models on ANUBIS 
and conduct an in-depth analysis of how different feature types affect recognition performance across 102 action categories. Our results show strong action-feature dependencies, highlight the limitations of na\"ive multi-representational fusion, and point toward the need for task-aware, semantically aligned integration strategies. This work offers both a comprehensive foundation and a practical benchmarking resource, aiming to guide the next generation of robust, generalizable skeleton-based action recognition systems for complex real-world scenarios.
The dataset, benchmarking framework, and code are available at \href{https://yliu1082.github.io/ANUBIS/}{https://yliu1082.github.io/ANUBIS/}.
\end{abstract}


\end{frontmatter}



\section{Introduction}

Human action recognition from visual data represents a fundamental challenge in computer vision, requiring robust extraction and analysis of spatiotemporal patterns from complex visual sequences~\cite{MINHDANG2020107561,  
lei_tip_2019}. The ability to automatically classify human activities from video data has enabled critical applications across intelligent surveillance systems~\cite{Park_2020_CVPR}, autonomous vehicle perception~\cite{kung2023action}, robotic interaction~\cite{Bahl_2023_CVPR}, augmented reality interfaces~\cite{Chatzitofis_2022_CVPR}, clinical motion analysis, and behavioral monitoring~\cite{10678592}.

To achieve robust action recognition, researchers have explored diverse data modalities \cite{lei_tip_2019}. RGB videos provide rich visual cues such as appearance, texture, and color, effectively capturing the overall dynamics of human movement and contextual information.
However, RGB-based methods face significant limitations: they suffer from performance degradation under challenging environmental conditions, including poor lighting, background clutter, and appearance variations, and as dense data, RGB videos are computationally intensive, typically requiring larger models and substantial computational resources for processing~\cite{Chen_2023_ICCV, Wasim_2023_ICCV, Yang_2022_CVPR}. These limitations collectively restrict their reliability and feasibility in real-world deployments. Depth maps offer complementary 3D structural information that enables better geometric and spatial modeling, yet depth-based methods remain sensitive to viewpoint changes, occlusions, and sensor noise \cite{Hu_2018_ECCV, lei_tip_2019}. 
Infrared data, while resilient to lighting variations, presents challenges for infrared-based methods, which often lack semantic richness and struggle to capture fine-grained motion details.
Given the limitations of these modalities, 3D human skeleton data, which represents human poses using a sparse set of key anatomical joints, has emerged as a highly effective representation for human action analysis \cite{2023_cvpr_3mformer,koniusz2021tensor}. 

\begin{figure}[tbp]
    \centering
    \includegraphics[width=\linewidth]{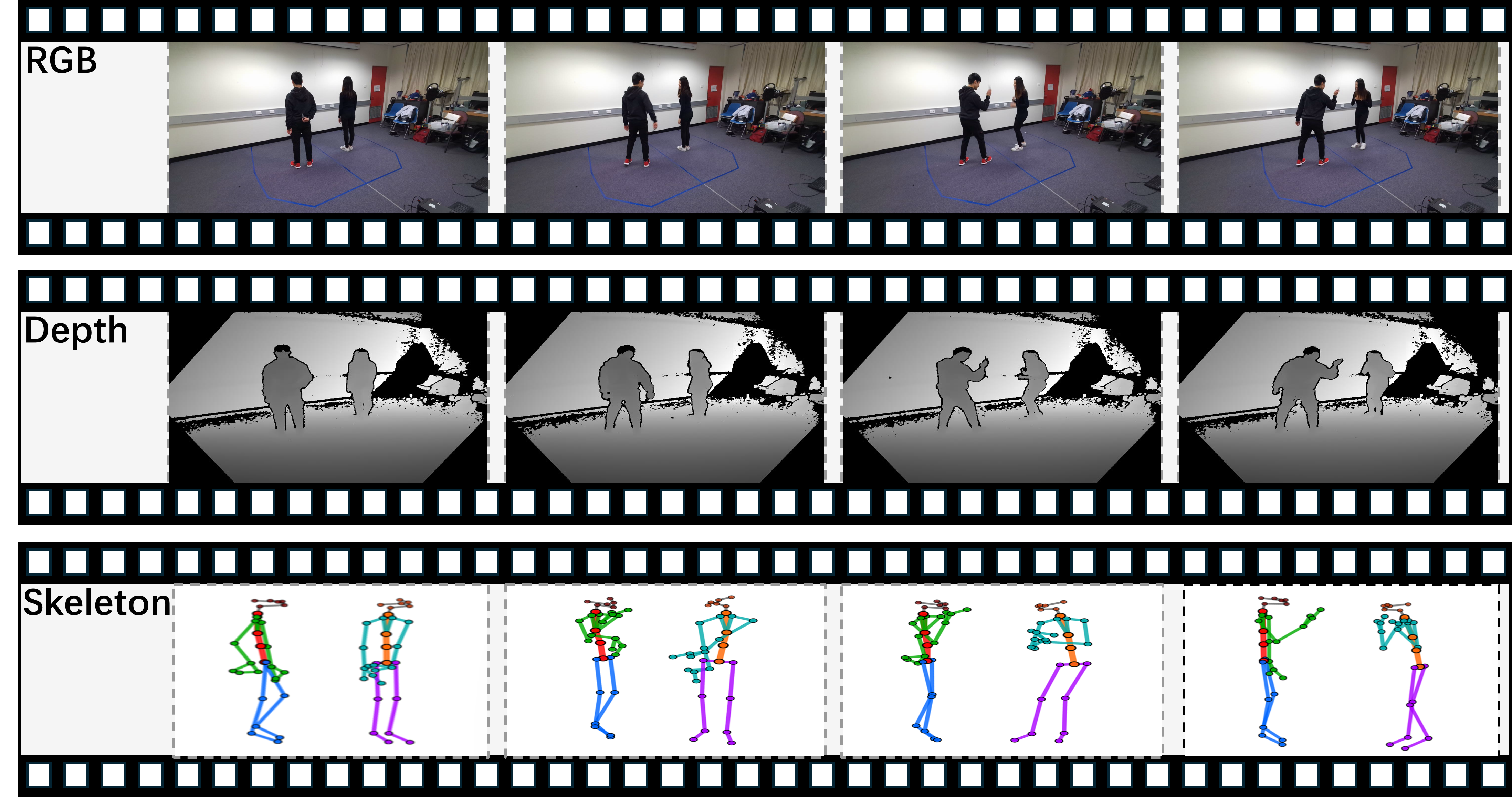}
    \caption{{Multi-modality example of the action \textit{wave knife to others} from the ANUBIS dataset, captured uniquely from a back-view perspective. Four consecutive frames are shown across RGB (top), Depth (middle), and 3D Skeleton (bottom) modalities. This example shows ANUBIS’s distinctive contributions in introducing previously unseen interaction classes and incorporating back-view acquisition, both absent in prior skeleton-based action datasets.}
}
    \label{fig:modality}
\end{figure}

Compared to other modalities, skeleton data offer several advantages (see Fig. \ref{fig:modality}): 
{reduced storage requirements} due to their sparse nature, computational efficiency during processing, robustness to environmental variations, invariance to appearance changes, and enhanced privacy protection by removing personally identifiable visual features. 
These qualities make skeleton data particularly well-suited for deployment in challenging real-world scenarios, edge devices with limited computational resources, and privacy-sensitive applications \cite{10678592,wangtaylor}. With recent advances in depth sensing technologies and pose estimation algorithms, 
acquiring high-quality skeleton sequences has become increasingly accessible.

{These advantages have driven significant advances 
across three key dimensions.}
Input representation has evolved beyond joint coordinates to incorporate bone vectors, velocities, accelerations, and surface normals for richer motion encoding~\cite{zhang2025generically}.
\textit{Spatial modeling} has been transformed by GCNs, with architectures like ST-GCN~\cite{2018_aaai_st_gcn}, 2s-AGCN~\cite{2019_cvpr_2sagcn}, and DeGCN~\cite{2024_tip_degcn} effectively capturing skeletal topology. 
\textit{Temporal modeling} has progressed from recurrent networks to advanced spatiotemporal convolutions (e.g., MS-G3D~\cite{2020_cvpr_msg3d}) and Transformers for long-range dependencies. 
Attention mechanisms, cross-modal fusion, and neural architecture search have further enhanced the performance. 
{As these directions continue to evolve, the choice of input representation becomes increasingly important, because different skeleton representations encode distinct geometric, structural, and temporal cues, and therefore often require different {spatiotemporal} modeling strategies. 
{Existing surveys have provided valuable overviews 
from broader perspectives \cite{sun2022human,kong2022human,xin2023transformer,liu2025systematic}}, including multimodal action understanding and general recognition/prediction tasks, comprehensive and systematic summaries of methods, datasets, applications, and challenges, as well as reviews focused on specific architectures or learning paradigms, such as Transformer-based action recognition and self-supervised skeleton representation learning. 
However, in these surveys, skeleton representation is typically discussed as one component within a wider review framework rather than being used as the central organizing principle. 
This motivates our representation-centric review, which uses input representation as the primary taxonomy and systematically relates joint, bone, motion, and derived skeleton representations to their corresponding {spatiotemporal} modeling designs.}

\begin{figure*}[tbp]
    \centering
    \includegraphics[width=\textwidth]{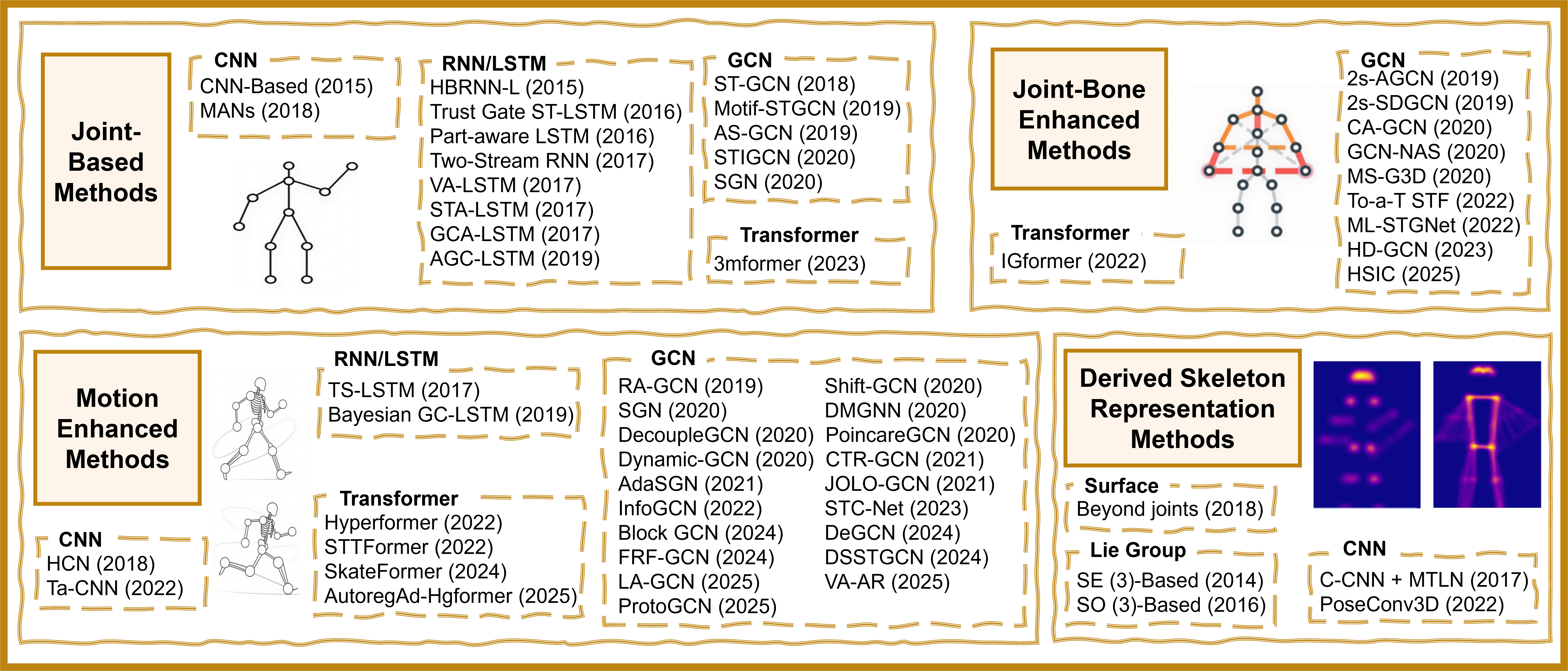}
    \caption{Evolution of {skeletal} action recognition methods from 2014 to 2025. The taxonomy categorizes approaches into four main groups: Joint Based (joint only), Joint-Bone Enhanced (joint+bone), Motion Enhanced (joint+bone+motion), and Derived Skeleton Representation Methods. Each category showcases the progression from traditional RNN/LSTM methods to modern GCN and Transformers, 
    demonstrating the evolution of deep learning techniques. 
    }
    \label{fig:arc}
\end{figure*}

{In addition to the need for a representation-centric synthesis, robust progress in this field also depends on challenging 
benchmark datasets.} 
Large-scale datasets, particularly the NTU RGB+D series~\cite{2016_cvpr_ntu,2019_tpami_ntu120}, have served as foundational benchmarks. 
However, these datasets increasingly fall short of meeting evolving research demands in several key aspects. First, the majority of actions are captured from frontal viewpoints with limited pose diversity, particularly \textit{lacking back-view perspectives from multiple angles}, which constrains model robustness when deployment scenarios involve varied camera angles or when subjects are oriented away from the primary sensor. Second, existing datasets \textit{predominantly focus on individual daily activities while neglecting complex multi-person interactions}, such as handshaking and collaborative behaviors, which are prevalent in real-world environments. Third, they \textit{fail to incorporate challenging actions involving aggression and violence} (e.g., hitting someone's head, stabbing with weapons), which represent important categories for security and surveillance applications but remain largely absent from current benchmarks. Finally, they \textit{lack contemporary socially relevant behaviors} (e.g., pandemic-related gestures and social distancing protocols), thereby limiting their ecological validity and applicability to modern scenarios. These gaps 
highlight the urgent need for more comprehensive datasets that better reflect the complexity and diversity of human actions in real-world.
{The main contributions of this work are as follows:
}
\renewcommand{\labelenumi}{\roman{enumi}.}
\begin{enumerate}
    \item 
    We provide a systematic taxonomy of skeleton-based action recognition methods {in the supervised setting}, organized by input representation (joints, bones, motion, extended features), and analyze how spatiotemporal modeling strategies adapt to each representation type.
    \item 
    We propose ANUBIS, a large-scale benchmark of 102 diverse actions. ANUBIS uniquely incorporates: (i) multiple viewpoints including back-view recordings, (ii) complex multi-person interactions, (iii) challenging violent and security-critical actions,
    filling critical gaps in existing datasets.
    \item 
    We evaluate a wide range of popular models on ANUBIS {under standardized supervised training/testing protocols}, showing how representation choice and modeling strategy affect performance, and uncovering cases where purely na\"ive multi-representational fusion degrades recognition. 
\end{enumerate}


\section{A Representation-Centric Review}
\label{SKELETON-BASED ACTION RECOGNITION: A REPRESENTATION-CENTRIC REVIEW}
\subsection{Joint-Based Methods}
Joint coordinates are characterized by multi-node structures, where the node count is typically determined by the skeletal detection sensors used~\cite{lei_tip_2019}. 
Each node encodes the spatial position of a human joint through either two-dimensional pixel coordinates or three-dimensional world coordinates, forming the geometric primitives of the skeletal graph structure. 
{These representations are inherently defined within Euclidean space, as they represent direct spatial locations of joints in either 2D or 3D Cartesian coordinates. This aligns with Euclidean geometry, where distances and angles between points are consistent, making joint coordinates ideal for capturing static positional relationships in the human body. In this context, the use of Euclidean space is advantageous for modeling the straightforward, rigid connections between joints that do not involve transformations or rotations.}

\subsubsection{Spatial Modeling}

Human pose analysis needs to capture both joint positions and the spatial relationships between them. Joint coordinates give the 2D/3D locations, while the skeletal structure comes from how joints connect and depend on each other. There are two main ways to model these spatial relationships: using pre-defined anatomical connections or learning the joint relationships 
from data.

\textbf{Predefined joint connectivity.}
Predefined connectivity methods embed fixed joint topological relationships directly into model architectures using binary adjacency matrices encoding anatomical connections. Unlike bone methods that add 3D geometric vectors as input features, predefined connectivity serves as structural constraints within the architecture. This approach offers computational efficiency since connectivity patterns remain static during training and inference, requiring negligible overhead.
Early CNNs use fixed skeletal connections within convolutions, where local receptive fields extract spatially-constrained features according to anatomical topology \cite{lei_tip_2019}. 
RNNs perform spatial modeling through sequence conversion using two strategies: sequential linearization transforms joint coordinates into structured orderings preserving local connectivity, while hierarchical partitioning uses two-layer architectures capturing local details and global integrity~\cite{2016_ECCV_Trust-Gate-ST-LSTM,2016_cvpr_ntu,2017_cvpr_gca_lstm,2017_cvpr_two_stream_rnn,2015_cvpr_hrnn,2017_iccv_va_lstm}. Beyond sequential approaches, GNNs use predefined adjacency matrices to constrain graph convolutions, enabling feature aggregation only among directly connected joints while constructing tree-like hierarchies based on parent-child relationships. These methods often convert absolute coordinates to local representations relative to parent nodes, eliminating global translation effects and incorporating geometric constraints for physiological plausibility~\cite{2018_aaai_st_gcn,2019_iccv_bayesian}.

\textbf{Hypergraph extensions for multi-joint coordination.}
Traditional binary-edge graphs prove insufficient for modeling complex multi-joint coordination patterns in human actions~\cite{2023_cvpr_3mformer}. 
Hypergraph extensions address this limitation through {
hyperedges that connect multiple joints simultaneously, enabling high-order relational modeling}~\cite{2021_TIP_hypergraph,2022_arxiv_hypergraph}. 
{Recent adaptive hypergraph methods further move beyond fixed or manually specified high-order structures. For example, Zhou et al.~\cite{zhou2025adaptive} introduce adaptive hyper-graph convolution with virtual connections, enabling the model to capture action-dependent coordination among joints that may not be directly connected by the physical skeleton. This provides a more flexible way to model high-order relational patterns than fixed pairwise adjacency or static hyperedge definitions.}
{Existing approaches also explore structured designs such as} hierarchical hyperedges: first-level edges maintain skeletal connections while higher-level edges aggregate multiple limb joints. 
This enables {multi-branch parallel processing frameworks that jointly capture} local motion features from low-order branches and coordination patterns from high-order branches via {unified matrix representations}~\cite{2023_arxic_dstgcn,2025_wacv_autoregressive}.


\textbf{Dynamic adaptive joint relationship learning.}
While manually predefined skeletal connections offer computational simplicity, they are limited to modeling adjacent joint relationships, constraining their ability to capture complex non-adjacent dependencies essential for sophisticated action recognition. To address this limitation, adaptive learning approaches have emerged where models automatically learn optimal connection patterns during training, enhancing spatial feature representation beyond fixed topological constraints.
Adaptive relationship learning provides two key advantages: (i) Enhanced spatial modeling enables learning both local adjacent relationships and global non-adjacent dependencies between arbitrary joint pairs, transcending physical connectivity limitations. For instance, clapping gestures require modeling cross-body relationships that extend beyond skeletal connections. (ii) Action-specific feature emphasis enables models to dynamically weight joint relationships based on discriminative importance for specific actions. In running motions, for example, lower limb joints carry greater semantic significance than upper limb joints, requiring adaptive emphasis on leg-related connections.

\textit{Attention-based adaptive weighting.} Attention mechanisms enable models to dynamically compute relationship strengths between joint pairs, learning fully-adaptive adjacency structures~\cite{2018_ijcai_man,2017_aaai_end_skeleton,2017_cvpr_gca_lstm,2018_eccv_sr_tsl,gedamu2023relation}. Widely deployed in recurrent architectures like RNNs and LSTMs, spatial attention computes functional associations between joints based on feature representations, generating data-driven connection weights across arbitrary spatial distances. This provides fully-adaptive topological modeling that autonomously discovers critical joint interactions across different motions.
Transformer-based architectures exemplify this approach through sophisticated self-attention mechanisms~\cite{2022_arxiv_sttformer, 2022_eccv_igformer,2024_ECCV_Skateformer,2022_arxiv_hypergraph,2025_wacv_autoregressive,2021_TIP_hypergraph, wangtaylor, li2026relposgar}. These models treat skeletal structures as fully-connected graphs where every joint can influence every other joint. Using Query-Key-Value mechanisms to calculate pairwise correlation scores, they dynamically determine the most relevant joint relationships for current input. While physical skeletal connectivity can be incorporated as positional bias, learned attention weights can deviate from anatomical constraints, enabling discovery of semantic relationships (e.g., hand-foot coordination) that transcend physical adjacency.

\textit{Specialized convolutional models.}
CNN-based approaches use channel-wise convolution to achieve global joint interaction modeling~\cite{2018_ijcai_hcn,2022_aaai_tacnn}. This paradigm represents skeletal sequences as 
tensors, then transposes them to a new tensor 
format where each joint becomes a distinct input channel. Subsequent $1 \!\times\! 1$ convolution across the 
channels implement learnable linear combinations of all joint features, effectively enabling each spatial location to aggregate information from all joints simultaneously. This channel-wise aggregation mechanism inherently captures global dependencies since each output feature incorporates weighted contributions from all input joints, regardless of their physical adjacency.

\textit{Learnable graph topology modification.}
GCN-based approaches use learnable parameters to modify connections within graph structures. These methods typically operate on predefined adjacency matrices encoding human skeletal topology, but augment them with trainable components for adaptive relationship learning.
Two primary strategies exist: (i) Adjacency matrix augmentation methods~\cite{2020_mm_stigcn,2019_aaai_motif_stgcn} learn additive or multiplicative modifications to fixed adjacency matrices, enabling models to strengthen existing connections or establish new pathways between previously unconnected joints. (ii) Learnable mask matrices share identical dimensionality with base adjacency matrices, where each trainable element modulates corresponding connection strength~\cite{2020_cvpr_sgn}. During training, these parameters adapt to emphasize discriminative joint relationships while potentially discovering cross-limb or non-adjacent dependencies complementing anatomical structure. Advanced variants~\cite{2019_cvpr_as_gcn} can learn sparse attention patterns that selectively activate non-adjacent connections based on action-specific requirements, effectively expanding receptive fields beyond immediate neighborhoods while preserving structural priors.
{Beyond adapting edges over a fixed skeleton, keypoint layout itself can also be refined. 
Yang et al.~\cite{yang2025expressive} introduce Expressive Keypoints by augmenting coarse body joints with selected hand and foot keypoints, and use Skeleton Transformation to reweight and progressively downsample the denser graph during GCN processing. 
This highlights keypoint granularity as another design axis for joint-based spatial modeling, balancing fine-grained expressiveness with computational efficiency.}

\subsubsection{Temporal Modeling}
Temporal modeling transforms static joint coordinates into dynamic motion trajectories by analyzing action boundaries, velocity variations, and sequential dependencies. This temporal analysis serves two critical functions: aggregating local motions across frames to form complete action semantics, resolving single-frame ambiguities; and enabling discrimination between spatially similar actions (e.g., putting on versus taking off clothing) through temporal pattern analysis \cite{wangtaylor}.
Temporal modeling faces inherent structural differences compared to spatial approaches \cite{qin2022strengthening}. Spatial modeling uses stable joint topology within individual frames, where joint relationships form consistent patterns reliably captured through single-frame analysis. In contrast, temporal modeling depends on inter-frame dynamics spanning multiple key frames, where action semantics emerge from continuous motion sequences rather than instantaneous spatial configurations.
This temporal dependency introduces three challenges. First, sampling-related issues arise from difficulty capturing complete action sequences: (i) Sparse frame sampling leads to incomplete action coverage, varying action rhythms cause semantic dilution, and short-duration movements are difficult to capture adequately. (ii) Fixed sampling strategies compound these problems, producing redundant frames for rapid motions and insufficient coverage for slower actions across different individuals and action types.
Second, robustness challenges emerge from noise and occlusion disturbances that propagate through sequential processing, creating information discontinuity, instability, and temporal imbalance that significantly constrain recognition performance~\cite{wangtaylor}. 
{This issue becomes more critical when skeleton sequences are temporally corrupted by missing frames or unreliable pose estimates. FineTec~\cite{shao2026finetec} addresses fine-grained action recognition under temporal corruption by combining sequence completion with skeleton decomposition, highlighting that robust temporal modeling should not only capture long-range dependencies but also recover corrupted motion cues and preserve subtle local dynamics.}
Third, computational challenges include long-range dependency attenuation, where local convolution operations dilute correlations between distant frames, and increased computational complexity from large receptive field requirements for complete action coverage.
Current approaches address these challenges through decomposed short-term and long-term modeling strategies for micro-actions and periodic sequences, respectively~\cite{2017_iccv_ts_lstm}. Effective solutions require preserving cross-frame joint correlations for spatiotemporal coherence, implementing adaptive time scales for motion variation accommodation, and emphasizing key frames for stable state learning. However, fundamental limitations persist in long-range dependency modeling and computational efficiency, necessitating continued methodological development.
Based on network structure, we categorize skeleton-based temporal methods into RNNs and spatiotemporal CNNs.

\textbf{RNNs} process skeleton sequences by combining current frames with historical memory states. Standard RNNs~\cite{2018_ijcai_man,2015_cvpr_hrnn} suffer from gradient issues limiting long-term modeling, making LSTMs~\cite{2016_ECCV_Trust-Gate-ST-LSTM,2016_cvpr_ntu,2017_cvpr_two_stream_rnn,2017_iccv_va_lstm,2017_aaai_end_skeleton,2017_cvpr_gca_lstm,2019_cvpr_agc_lstm} more suitable.
LSTM improvements focus on hierarchical modeling and global context integration. Hierarchical methods partition joints into body parts, using part-specific LSTMs for local dynamics and higher-level LSTMs for inter-part relationships~\cite{2017_iccv_ts_lstm,2016_cvpr_ntu,2017_cvpr_two_stream_rnn}. Global context approaches use temporal attention to dynamically weight time steps and integrate historical features via context vectors~\cite{2017_aaai_end_skeleton,2017_cvpr_gca_lstm}. These enhancements improve dependency modeling. 

\textbf{{Spatiotemporal} CNNs.}
Temporal convolution applies one-dimensional kernels along the time axis to aggregate adjacent frame features, using dilated convolution to expand receptive fields for long-range dependency capture~\cite{
2017_cvpr-w_Res-TCN}. However, pure temporal convolution ignores spatial joint structure and requires deep networks or large dilation rates for adequate receptive fields, increasing computational costs. Research has shifted toward integrated spatiotemporal convolution that jointly models spatial correlations and temporal dynamics~\cite{2019_aaai_motif_stgcn,2018_aaai_st_gcn,2020_cvpr_sgn,2020_mm_stigcn,2019_cvpr_as_gcn}.
Subsequent improvements focus on computational efficiency, representation enhancement, and more flexible spatiotemporal topology modeling. Residual connections optimize gradient propagation, while temporal attention mechanisms guide focus toward critical periods~\cite{2018_ijcai_man,2025_aaai_va-ar,2020_mm_stigcn}. {Recent studies further improve spatiotemporal graph modeling by introducing frequency-aware topology learning~\cite{xia2026frequency} and continual spatiotemporal graph convolutional learning~\cite{hedegaard2023continual}.} Multi-scale temporal aggregation through dilated convolution captures dependencies at varying time scales, integrating short-term and long-term segments for complex action modeling.

Neural architecture search 
enables automated spatiotemporal convolution optimization~\cite{2020_aaai_gcn_nas}. By defining search spaces with temporal kernel sizes, dilation rates, and connectivity patterns, evolutionary algorithms and reinforcement learning discover efficient architectures without manual design biases, achieving optimal hierarchical temporal modeling.
Non-Euclidean temporal embedding provides alternative modeling approaches~\cite{2020_cvpr_msg3d, 2020_cvpr_shift_gcn,2019_aaai_motif_stgcn}. Hyperbolic space mapping {uses} exponential distance growth to enhance long-range temporal dependency discrimination. This embeds temporal dynamics into curved spaces via logarithmic and exponential transformations, reducing distant feature confusion in Euclidean spaces, 
valuable for actions requiring subtle temporal distinctions.

{
Joint-based methods remain the most fundamental representation for skeleton-based action recognition, but their effectiveness depends on how well models capture adaptive spatial and temporal dependencies from sparse joint coordinates. Future designs should move beyond fixed anatomical topology and develop action-aware joint relationship learning, robust key-frame selection, and long-range temporal modeling, especially for viewpoint changes, occlusions, and complex multi-person interactions.
}

\subsection{Joint-Bone Enhanced Methods}
\label{Joint-Bone Enhanced Methods}
Joint-bone enhanced methods explicitly generate bone vectors by computing coordinate differences between joints during preprocessing, resulting in independent input features that encode limb length, direction, and anatomical connectivity~\cite{2019_cvpr_2sagcn,2019_cvpr_dgnn}. {These bone vectors represent the relative spatial relationships between joints, capturing key attributes such as the limb's length and orientation. By encoding the direction and distance between connected joints, bone vectors provide a direct representation of the body's skeletal structure, which is particularly useful for modeling human motion. Unlike joint-based methods, which focus solely on individual joint positions, joint-bone methods enhance the feature set by incorporating the structural relationships between joints, improving action recognition performance.}
Spatial modeling with bone explicitly incorporates anatomical structural information as additional input alongside joint coordinates. This approach computes bone vectors during preprocessing and feeds them as independent streams providing explicit joint relationships. Bone vectors are calculated as directional vectors between anatomically connected joints~\cite{wangtaylor}.
Existing joint-bone methods fall into two paradigms: dual-stream architecture and information fusion architecture.

\textbf{Dual-stream architecture} uses independent two-stream networks to process joint and bone representations separately, with integration occurring at the decision level through weighted combination of prediction scores \cite{ 2019_cvprw_2s_sdgcn,2022_aaai_to-a-t,2022_tip_ML-STGNet,2018_ijcai_hcn,2020_aaai_gcn_nas,2022_aaai_tacnn,2024_CVPR_BlockGCN,2022_CVPR_infogcn,2023_iccv_hdgcn}. 
Representative methods include 2s-AGCN~\cite{2019_cvpr_2sagcn}, MSG3D~\cite{2020_cvpr_msg3d}, and Dynamic GCN~\cite{2020_mm_dynamic_gcn}.
While this approach ensures feature-specific learning without cross-modal interference, it requires multiple times the parameters and computation of single-stream methods. Additionally, late fusion may miss intricate inter-modal dependencies, potentially underutilizing complementary information between joint coordinates and bone structural features.

\textbf{Feature fusion architecture} {integrates} bone as auxiliary contextual features within a single framework, enhancing joint representations through early or intermediate fusion~\cite{2022_eccv_igformer,2025_wacv_autoregressive,2019_cvpr_dgnn}. CA-GCN~\cite{2020_cvpr_ca_gcn} incorporates bone information via context-aware mechanisms that compute attention weights to determine bone feature relevance for each joint. This context term enriches the primary joint stream with structural skeleton information.
Unlike dual-stream methods, this paradigm uses a single network processing the primary joint stream while adding bone as supplementary context, achieving significantly lower parameters and computational cost. However, unified processing may cause feature interference where different characteristics lead to suboptimal joint representations~\cite{wang2019hallucinating}, and the single network may not fully exploit unique properties that dedicated streams could capture.

Spatial modeling approaches for joint-bone representations fundamentally differ from joint-only methods, including manually predefined joint connectivity and dynamic adaptive joint relationship learning, which construct bone relationships implicitly within networks. Explicit bone preprocessing offers distinct advantages: pre-computed bone vectors directly encode physical connections between adjacent and non-adjacent joints, using anatomical prior knowledge for immediate access to limb relationships and biomechanical constraints. This design concentrates computational costs in preprocessing while requiring only multi-stream feature fusion during inference, contrasting with joint-only methods that {need to} predefine or learn these relationships within networks. Each method presents characteristic trade-offs. Joint-bone enhanced methods provide computational efficiency and anatomical consistency but are limited to predefined structural relationships. Among joint-only approaches, manually predefined connectivity offers simplicity but restricts modeling to adjacent joints, while dynamic adaptive learning enables flexible capture of non-adjacent joint relationships, a key advantage for complex action modeling, but requires large-scale training data and incurs higher computational costs. The choice depends on application requirements: joint-bone enhanced methods suit scenarios requiring anatomical consistency and efficiency, while dynamic adaptive approaches excel when flexible non-adjacent joint modeling is crucial for complex actions.

{
Joint-bone enhanced methods offer strong anatomical priors and efficient structural encoding, but fixed bone definitions may introduce irrelevant or redundant cues for some actions. Future designs should explore adaptive bone selection, task-aware bone weighting, and more flexible joint-bone interaction modeling, especially for fine-grained actions and multi-person scenarios where subtle limb coordination is critical.
}

\subsection{Motion Enhanced Methods}
Motion represents dynamic changes extracted from temporal variations between adjacent frames, capturing kinematic dependencies and dynamic spatial information essential for distinguishing actions that remain ambiguous through spatial analysis alone. {These dynamic features can be classified into two main categories: joint motion information and bone motion information.}

\textbf{Joint motion information} captures the temporal changes of individual joints, typically represented as coordinate differences between consecutive frames. For each joint, its motion feature is defined as the change in position relative to the previous frame, effectively capturing joint velocity \cite{wangtaylor}. {This representation focuses on the temporal evolution of each joint’s position, making it suitable for detecting motions where individual joint movement is significant, such as walking or waving.}

\textbf{Bone motion information} captures the temporal changes in the relationships between connected joints, reflecting how the relative positions of connected joints evolve from one frame to the next. This representation is 
important for modeling actions that involve complex structural changes, such as limb extensions or coordinated movements across multiple limbs, as it emphasizes skeletal connectivity dynamics rather than simply focusing on individual joint displacement~\cite{2019_cvpr_dgnn}. {By encoding the movement between multiple joints simultaneously, bone motion information offers a more holistic view of the body's motion, helping to capture actions that require coordination, like jumping or throwing.}

Recent research adopts two main architectures for motion integration. Multi-stream models process joint, bone, and motion features in separate branches, then fuse them via concatenation or weighted fusion \cite{2020_cvpr_shift_gcn,2020_mm_dynamic_gcn,2021_iccv_ctr_gcn,2021_iccv_adasgn,2023_iccv_stc-net,2025_cvpr_porotogcn,2024_TIP_DSSTGCN,2025_TMM_lagcn}. In contrast, unified approaches directly integrate features through channel-wise concatenation, while motion streams often incorporate refined temporal modeling to capture dynamics \cite{2019_icip_ra_gcn,
2020_cvpr_sgn,2020_cvpr_ca_gcn,2025_wacv_autoregressive,2020_cvpr_dmgnn,2024_aaai_frf-gcn}.
Both architectures benefit from two key strategies. Attention mechanisms focus on informative motion components \cite{2020_cvpr_sgn,2022_CVPR_infogcn,2025_aaai_va-ar,2025_cvpr_porotogcn,2024_tip_degcn}: spatial attention highlights active joints, temporal attention identifies keyframes, and cross-modal attention captures complementary relationships. Multi-scale temporal modeling captures motion at varying time scales \cite{2022_CVPR_infogcn,2024_CVPR_BlockGCN} using parallel convolution branches: small kernels for fast motions and large kernels for long-duration actions.

{
Motion-enhanced methods are particularly effective when action semantics depend on temporal evolution, such as cyclic movements, phase transitions, or subtle velocity changes. However, motion cues are also sensitive to frame sampling, pose estimation noise, and action-speed variation, which may introduce unstable or redundant information when the target action is mainly defined by static posture or global spatial configuration. Future work should therefore develop phase-aware motion modeling, adaptive temporal scale selection, and reliability-aware motion fusion, so that motion features are emphasized only when they provide discriminative temporal evidence.
}

\subsection{Derived Skeleton Representation Methods}
Researchers explore derived skeleton representations with enhanced expressiveness through mathematical transformations, physical model derivations, or visual processing techniques.

\textbf{Surface normal} approaches represent body part geometry by capturing relative shapes from adjacent edges. Each connected edge pair has a surface normal vector describing the corresponding body surface orientation. To match joint coordinate scales, these vectors are appropriately scaled. For multi-joint skeletons, each joint defines a surface, creating a tensor with dimensions for frames, joints, and the three normal vector components~\cite{2018_TIP_surface}.
Surface normals capture shape information during movements, including planar angle changes from body parts, enhancing multi-joint spatial interaction representation and limb posture modeling. In temporal modeling, surface normal variations represent planar configuration evolution in limb movements, complementing joint and edge features. Bidirectional LSTM processing enhances time-dependent action learning.
However, surface normal computation has two limitations. First, edge vector calculation errors from noise or occlusion propagate to surface normal vectors, introducing erroneous geometric dynamics. Second, retaining only planes may lose geometric correlations between non-adjacent joints in time series, affecting complete expression of subtle temporal features.

{\textbf{Lie group-based} methods} represent human skeletons using two frameworks{: (i)} SE(3)-based approaches model skeletons as rigid body transformations, capturing rotation and translation to describe spatial pose relationships~\cite{2014_cvpr_lie}{, and (ii)} SO(3)-based approaches focus on relative rotational relationships between body parts, using only rotation matrices with scale normalization to retain 3D rotational information. This creates scale-invariant skeletal representation while emphasizing rotational dynamics~\cite{2016_cvpr_Rolling}.
Both SE(3)- and SO(3)-based methods model action sequences as curves on Lie group manifolds using Dynamic Time Warping (DTW) for temporal variations. SE(3) methods project action curves to Lie algebra, extract multi-scale features via Fourier Time Pyramid, and classify using linear SVM, capturing patterns at different temporal resolutions. SO(3) methods use rolling mapping to unfold manifolds along action curves, reducing distortion while preserving distance relationships for accurate rotational modeling. Lie group representations provide geometrically meaningful skeleton modeling: SE(3) methods preserve complete pose information (rotation and translation), while SO(3) methods achieve efficiency and cross-individual generalization through rotational focus.

However, these approaches have several limitations. SE(3) representations incur higher computational complexity due to increased feature dimensionality. SO(3) methods may sacrifice spatial translation information relevant for certain actions. Both show decreased discriminative ability with extreme postures or complex action couplings due to nonlinear manifold characteristics. Additionally, rolling mapping algorithms, while reducing distortion, introduce implementation complexity that may limit practical deployment \cite{lei_tip_2019}.

\textbf{Joint heatmap} methods convert 2D skeleton coordinates into probabilistic spatial representations using Gaussian distributions. Each joint becomes a heatmap reflecting location and confidence, with Gaussian spread controlling positional uncertainty. Stacking heatmaps over time creates a 3D volume capturing spatial and temporal skeleton information with dimensions for joints, temporal length, and spatial resolution~\cite{2022_cvpr_poseconv3D}.
The probabilistic representation enables robust pose estimation under noise while maintaining efficiency through direct multi-joint accumulation. The 3D volume structure facilitates direct 3D-CNN processing for spatiotemporal feature extraction.
%
\textit{Spatial modeling} uses adapted 3D-CNN architectures that remove early downsampling to preserve low-resolution heatmap features. Shallow networks reduce complexity while maintaining spatial dependency capture. Preprocessing involves bounding box localization and cropping to focus on subjects while preserving relationships. Multi-channel input combining joint and limb heatmaps enhances structure representation.
\textit{Temporal modeling} uses 3D convolution kernels to simultaneously 
extract spatial joint relationships and temporal changes across time steps. This unified processing captures dynamic joint evolution patterns. 
Heatmap representations provide noise robustness through probabilistic encoding and efficient 3D-CNN compatibility, enabling direct spatial modeling without explicit graph construction while maintaining tractability.
However, limitations include spatial resolution constraints that may lose fine-grained information, increased memory requirements for 3D volumes, and potential temporal redundancy in slow-varying actions. The Gaussian assumption may not optimally represent all joint confidence distributions, particularly for occluded or uncertain detections.

{
Derived skeleton representations show that 
representation can 
reshape what information is visible to the model. Unlike joint, bone, or motion features, which directly describe body state or temporal change, derived representations impose additional geometric or probabilistic structures on skeleton sequences, such as surface orientation, manifold constraints, or image-like spatial distributions. This can improve discriminability when the imposed structure matches the action, but may also introduce unnecessary bias when it does not.
}

\section{A New Benchmark Dataset}
\label{sec:dataset}


\textbf{Motivation.} ANUBIS dataset addresses three key limitations. 
First, established datasets like NTU60~\cite{2016_cvpr_ntu} and NTU120~\cite{2019_tpami_ntu120} capture primarily frontal and lateral viewpoints while excluding rear-view perspectives, limiting robustness for challenging observation positions common in real-world monitoring. Second, these benchmarks focus on individual behaviors while underrepresenting multi-person interactions like collaborative activities and assistance behaviors prevalent in practical applications. Third, existing datasets show constrained action scope, omitting aggressive behaviors critical for security surveillance (stabbing, striking) and pandemic-induced social adaptations (elbow touching, arm-directed sneezing) that emerged post-COVID-19. These gaps reduce real-world utility, as models cannot recognize actions from challenging angles, multi-person interactions, and new behaviors essential for practical systems.

ANUBIS offers distinct advantages over existing datasets in three key aspects:
{\textit{(i) Technical specifications.}} We use Microsoft Azure Kinect devices, {which provide synchronized RGB, depth, and skeleton acquisition with more reliable skeleton tracking than 
Kinect V1/V2 systems used in previous benchmarks.} Compared with NTU's 25-joint format, ANUBIS captures 32 joint coordinates per skeleton, {including additional facial and clavicle joints that support richer upper-body and interaction analysis.}
{\textit{(ii) Content scope and diversity.}} ANUBIS addresses critical gaps by expanding action categories. While NTU-60 (60 categories, 40 participants, 56,880 sequences) and NTU-120 (120 categories, 106 participants, 114,480 sequences) emphasized scale expansion, ANUBIS (102 categories, 80 participants, 66,232 clips) introduces qualitative advances through three previously underrepresented types: complex multi-person interactions, aggressive behaviors for security applications, and contemporary pandemic-induced social adaptations.
{\textit{(iii) Comprehensive viewpoint coverage.}} The distinctive innovation lies in systematic rear-view data collection across multiple angles. This creates challenges where hand details and frontal movements become occluded, better approximating real-world monitoring where subjects may not face cameras directly. This balanced frontal-posterior perspective distribution addresses observational limitations in traditional datasets while enhancing model robustness under diverse camera orientations. 

{
To better clarify the dataset components, Fig.~\ref{fig:modality} provides a representative ANUBIS example with synchronized RGB, depth, and 3D skeleton sequences. ANUBIS contains 102 action categories collected from 80 participants, producing 66,232 skeleton clips across diverse viewpoints. Each clip includes three modalities: RGB video, depth video, and 3D skeleton data with 32 joints per person. For benchmarking, we trained with the front view and test on the back view.
All evaluated methods follow the same split and preprocessing protocol to ensure fair comparison. More details on action design, camera setup, preprocessing, and statistics are provided in the Appendix.
}

\begin{table*}[tbp]
\setlength{\tabcolsep}{0.15em}
\renewcommand{\arraystretch}{0.70}
\centering
\small
\caption{{Performance of representative 
on NTU60, NTU120, and the proposed ANUBIS. While NTU60 and NTU120 results are approaching saturation, ANUBIS presents a substantially more challenging and unsolved benchmark, leaving significant room for performance improvement. }The $\textcolor{best}{\textbf{best}}$ and $\textcolor{second}{\textbf{second-best}}$ performances are highlighted.}
\label{performance}
\resizebox{0.9\textwidth}{!}{
\begin{tabular}{l c c c c c c c c c c c c}
\toprule
\multirow{3}{*}{\textbf{Method}} & \multirow{3}{*}{Venue} & \multicolumn{3}{c}{\textbf{Features}} & \multicolumn{6}{c}{\textbf{Dataset}} & \multirow{3}{*}{\shortstack{\textbf{Params}\\\textbf{(M)}}} & \multirow{3}{*}{\textbf{GFLOPs}} \\
\cmidrule(lr){3-5} \cmidrule(lr){6-11}
 &  & \textbf{J} & \textbf{B} & \textbf{M} & \multicolumn{2}{c}{\textbf{NTU60}} & \multicolumn{2}{c}{\textbf{NTU120}} & \multicolumn{2}{c}{\textbf{ANUBIS}} &  &  \\
\cmidrule(lr){6-7} \cmidrule(lr){8-9} \cmidrule(lr){10-11}
 &  &  &  &  & \textbf{X-Sub} & \textbf{X-View} & \textbf{X-Sub} & \textbf{X-Set} & \textbf{Top-1} & \textbf{Top-5} &  &  \\
\midrule
STGCN\cite{2018_aaai_st_gcn} & AAAI'18 & \checkmark & & & 81.5 & 88.3 & - & - & 50.25 & 79.96 & 3.4 & 45.23 \\
Motif-STGCN\cite{2019_aaai_motif_stgcn} & AAAI'19 & \checkmark & & & 84.2 & 90.2 &- & - & 55.76 & 83.96 & 1.78 & 27.10 \\
\midrule[0.3pt]
2s-AGCN\cite{2019_cvpr_2sagcn} & CVPR'19 & \checkmark & \checkmark & & 88.5 & 95.1 &- & - & 57.26 & 84.86 & 3.47 & 47.84 \\
MS-G3D\cite{2020_cvpr_msg3d} & CVPR'20 & \checkmark & \checkmark & & 91.5 & 96.2 & 86.9 & 88.4 & 54.17 & 82.05 & 3.8 & 62.72 \\
GCN-NAS\cite{2020_aaai_gcn_nas} & AAAI'20 & \checkmark & \checkmark & & 89.4 & 95.7 &- & - & 56.40 & 84.37 & 6.57 & 93.64 \\
HD-GCN\cite{2023_iccv_hdgcn} & ICCV'23 & \checkmark & \checkmark & & 93.4 & 97.2 & 90.1 & 91.6 & 51.33 & 80.96 & 8.8 & 12.74 \\
\midrule[0.3pt]
RA-GCN\cite{2019_icip_ra_gcn} & ICIP'19 & \checkmark & & \checkmark & 85.9 & 93.5 &- & - & 41.87 & 73.45 & 10.26 & 135.52 \\
\midrule[0.3pt]
Shift-GCN\cite{2020_cvpr_shift_gcn} & CVPR'20 & \checkmark & \checkmark & \checkmark & 90.7 & 96.5 & 85.9 & 87.6 & 26.84 & 57.50 & 0.73 & 6.16 \\
Decoupling-GCN\cite{2020_eccv_decouple_gcn} & ECCV'20 & \checkmark & \checkmark & \checkmark & 90.8 & 96.6 & 86.5 & 88.1 & 52.32 & 80.06 & 3.63 & 32.98 \\
CTR-GCN\cite{2021_iccv_ctr_gcn} & ICCV'21 & \checkmark & \checkmark & \checkmark & 92.4 & 96.8 & 88.9 & 90.4 & 37.90 & 70.40 & 10.07 & 141.68 \\
STTFormer\cite{2022_arxiv_sttformer} & arXiv'22 & \checkmark & \checkmark & \checkmark & 92.3 & 96.5 & 88.3 & 89.2 & 57.77 & 85.30 & 6.6 & 109.08 \\
Hyperformer\cite{2022_arxiv_hypergraph} & arXiv'22 & \checkmark & \checkmark & \checkmark & 92.9 & 96.5 & 89.9 & 91.3 & 47.51 & 77.77 & 3.1 & 32.68 \\
InfoGCN\cite{2022_CVPR_infogcn} & CVPR'22 & \checkmark & \checkmark & \checkmark & 93.0 & 97.1 & 89.4 & 90.7 & 46.99 & 76.69 & 1.6 & 19.97 \\
BlockGCN\cite{2024_CVPR_BlockGCN} & CVPR'24 & \checkmark & \checkmark & \checkmark & 93.1 & 97.0 & 90.3 & 91.5 & 54.46 & 81.73 & 2.5 & 36.79 \\
Skateformer\cite{2024_ECCV_Skateformer} & ECCV'24 & \checkmark & \checkmark & \checkmark & \textcolor{second}{\textbf{93.5}} & \textcolor{best}{\textbf{97.8}} & 89.8 & 91.4 & 45.02 & 75.43 & 3.8 & 8.93 \\
DS-STGCN\cite{2024_TIP_DSSTGCN} & TIP'24 & \checkmark & \checkmark & \checkmark & 93.2 & \textcolor{second}{\textbf{97.5}} & 89.4 & 91.2 & 52.43 & 81.96 & 1.4 & 14.09 \\
DeGCN\cite{2024_tip_degcn} & TIP'24 & \checkmark & \checkmark & \checkmark & \textcolor{best}{\textbf{93.6}} & 97.4 & \textcolor{best}{\textbf{91.0}} & \textcolor{best}{\textbf{92.1}} & \textcolor{second}{\textbf{60.16}} & \textcolor{second}{\textbf{85.63}} & 1.4 & 9.75 \\
ProtoGCN\cite{2025_cvpr_porotogcn} & CVPR'25 & \checkmark & \checkmark & \checkmark & \textcolor{second}{\textbf{93.5}} & \textcolor{second}{\textbf{97.5}} & 90.4 & \textcolor{second}{\textbf{91.9}} & 47.56 & 78.10 & 4.2 & 29.88 \\
LA-GCN\cite{2025_TMM_lagcn} & TMM'25 & \checkmark & \checkmark & \checkmark & \textcolor{second}{\textbf{93.5}} & 97.2 & \textcolor{second}{\textbf{90.7}} & 91.8 & \textcolor{best}{\textbf{60.33}} & \textcolor{best}{\textbf{86.87}} & 3.4 & 28.32 \\
\bottomrule
\end{tabular}} 
\end{table*}

\section{Evaluation and Benchmarking}


{Table~\ref{performance} presents the benchmark comparison.}


\textbf{Performance saturation on NTU.}
Results on NTU60 and NTU120 show clear performance saturation. Recent methods, including DeGCN, LA-GCN, and ProtoGCN, exceed 93\% Top-1 accuracy on NTU60 cross-subject and over 97\% on cross-view splits. On NTU120, these methods regularly surpass 90\%, leaving minimal improvement headroom. This plateau suggests NTU datasets, while historically instrumental, no longer fully differentiate state-of-the-art skeleton-based recognition capabilities. Consequently, incremental NTU gains may not reflect real-world robustness or generalization ability.

\textbf{ANUBIS as a more challenging benchmark.}
In contrast, ANUBIS results are substantially lower, even for top-performing models. LA-GCN achieves the highest Top-1 accuracy at 60.33\%, closely followed by DeGCN at 60.16\%, with Top-5 accuracies just above 85\%. Most other methods fall in the 40-55\% range, despite reaching near-perfect scores on NTU. This performance drop highlights ANUBIS's increased difficulty, driven by rear-view occlusions, fine-grained actions, and modern social interaction patterns. The dataset's complexity disrupts conventional representation exploitation patterns, indicating that current architectures, optimized for standard benchmarks, struggle to adapt to more realistic and diverse action scenarios.

\textbf{Representation-type 
variability.}
Interestingly, ANUBIS does not show straightforward correlation between input feature types and performance. Joint-only (J) methods like Motif-STGCN reach 55.76\%, surpassing some tri-representation (Joint+Bone+Motion) methods such as CTR-GCN (37.90\%) and Shift-GCN (26.84\%). Even within tri-representation designs, variance is high: LA-GCN and DeGCN lead with 60\%+ Top-1, while others hover below 50\%. This suggests feature fusion quality and architectural adaptability outweigh simply increasing feature types. On ANUBIS, fusion design, attention mechanisms, and temporal-spatial reasoning appear more critical than multiple feature presence.

\textbf{Computational efficiency considerations.}
From a resource perspective, there is no consistent accuracy-efficiency trade-off. Lightweight models like DeGCN (1.4M parameters, 9.75 GFLOPs) achieve top-tier performance, while complex models such as CTR-GCN (10.07M parameters, 141.68 GFLOPs) perform significantly worse on ANUBIS. This efficiency-performance decoupling shows computational cost does not guarantee robustness on challenging datasets. It also highlights potential for resource-friendly yet highly accurate architectures, particularly for edge and real-time deployment scenarios.

\begin{figure*}[tbp]
    \centering
    \includegraphics[width=0.95\textwidth]{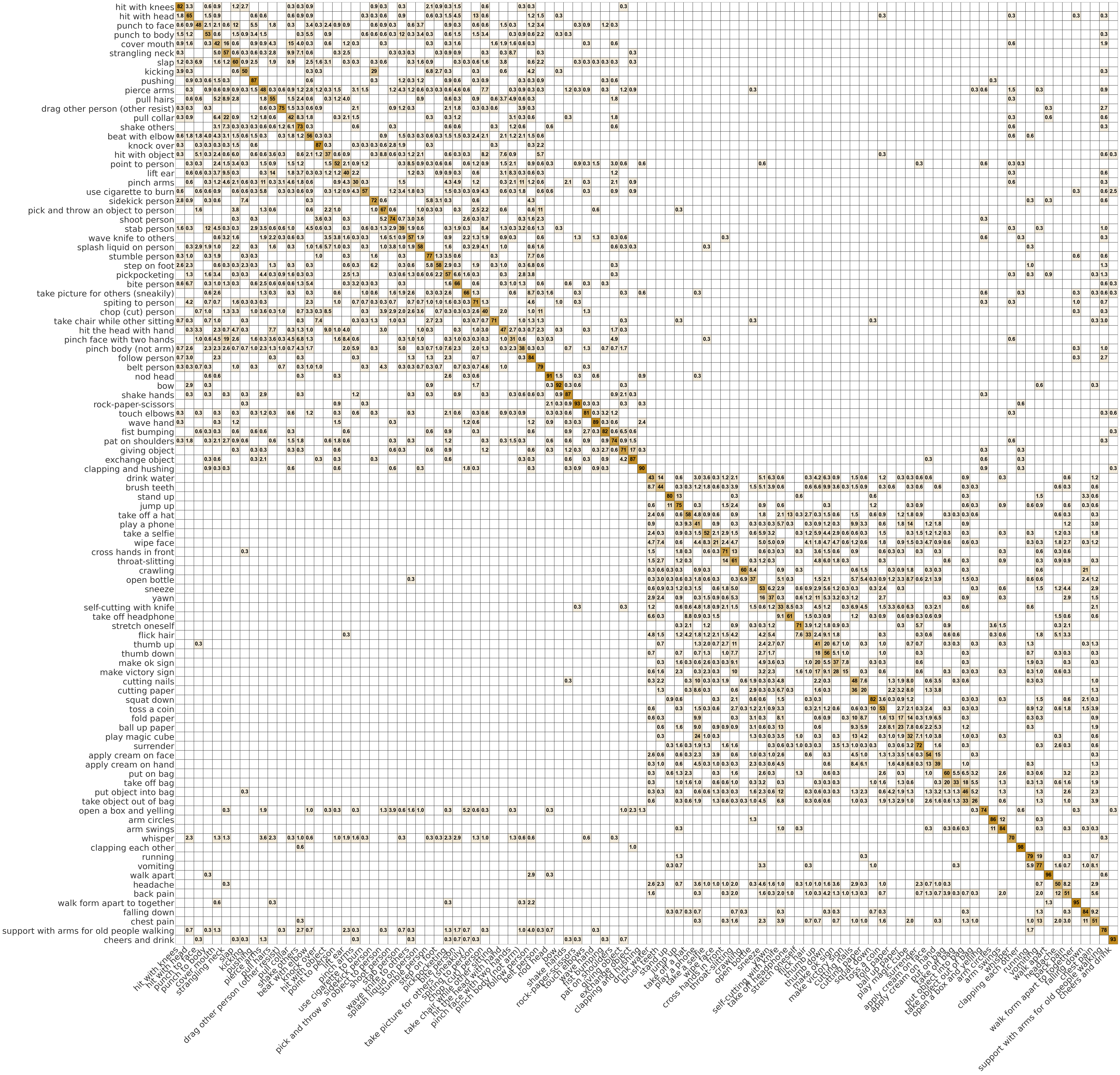}
    \caption{Confusion matrix of LA-GCN on ANUBIS. Clear diagonal dominance reflects strong recognition for macro-motion and dyadic actions, while off-diagonal confusions cluster among fine-grained, hand/object-centric classes (e.g., fold paper, play magic cube), highlighting the limitations of skeleton-only features in capturing subtle manipulations and object context.}
    \label{fig:confusion_matrix}
\end{figure*}

\section{In-Depth Analysis and Discussion}

Below, we provide an in-depth analysis of the ANUBIS dataset evaluations.

\subsection{Confusion Matrix Analysis and Dataset Challenges}

We present an in-depth analysis of  confusion matrix for ANUBIS using LA-GCN (best model), shown in Fig. \ref{fig:confusion_matrix}.
The confusion matrix for LA-GCN on ANUBIS shows highly non-uniform recognition accuracy across action classes. Certain categories with distinctive, large-scale movements like clapping each other, walk apart, and walk from apart to together form bright diagonal clusters, indicating high accuracy with minimal confusion. These actions provide strong spatiotemporal signals and distinctive inter-joint relationships that LA-GCN models effectively. In contrast, many fine-grained or localized actions display heavy off-diagonal activity, reflecting substantial misclassification. The imbalance between well-separated macro-movements and ambiguous micro-movements underscores the diverse difficulty spectrum.

\textbf{Sources of misclassification.}
High confusion rates are particularly evident among actions with similar skeletal motion trajectories but differing in object interactions or subtle hand gestures, for example, fold paper vs. cutting paper, make victory sign vs. thumb up, or apply cream on face vs. wash face. Since ANUBIS is purely skeleton-based, critical cues such as object presence, fine finger articulation, or texture changes are absent, forcing the model to infer action semantics from incomplete spatiotemporal data. Additionally, rear-view sequences exacerbate this challenge, as many hand and torso movements become partially occluded, reducing the discriminability of similar gestures.

\textbf{Viewpoint and interaction complexity.}
ANUBIS's rear-view and multi-person scenarios present significant challenges. Rear-view cases occlude key joints (hands, face), increasing confusion between distinct actions. Interactive actions sometimes benefit from relational cues (proximity, coordination), outperforming isolated fine-motor actions. However, overlapping motions cause inter-person occlusions and role ambiguity, increasing errors between cooperative and antagonistic interactions.

\textbf{Why ANUBIS is a valuable benchmark?}
This confusion matrix demonstrates ANUBIS's substantially greater difficulty compared to traditional benchmarks like NTU. It combines viewpoint variation, fine-grained gestures, modern social behaviors, and complex multi-person interactions, challenging models' spatial reasoning, temporal modeling, and robustness to incomplete data.
ANUBIS provides opportunities for future research: (i) Multi-modal fusion: integrating RGB, depth, or object-context streams to resolve semantic ambiguities. (ii) Fine-grained feature learning: improving finger joint precision and hand region attention. (iii) Viewpoint-invariant representations: developing architectures resilient to occlusions and perspective shifts. (iv) Relational reasoning: enhancing inter-person dynamics modeling for cooperative and competitive interactions.

\textbf{Implications for model design.}
The persistent confusions highlighted here suggest that future algorithms should go beyond current GCN and Transformer hybrids, which excel in macro-motion capture but falter in local detail extraction. Promising directions include hierarchical feature learning, where models capture both global body structure and localized joint dynamics; self-supervised pretraining to enrich motion semantics without additional labels; and scene- or object-aware skeleton augmentation to inject missing contextual signals. By tackling the challenges exposed by ANUBIS, researchers can develop models with stronger generalization, adaptability to real-world occlusions, and robustness to nuanced human behaviors, capabilities essential for next-generation action understanding systems.

\begin{table*}[tbp]
\centering
\setlength{\tabcolsep}{0.2em}
\renewcommand{\arraystretch}{0.7}
\caption{{Per-class accuracy of six leading models on ANUBIS, showing the top-20 easiest and bottom-20 most challenging actions.
{Model abbreviations: LA (LA-GCN), STT (STTFormer), 2s (2s-AGCN), ST (STGCN), M-ST (Motif-STGCN), De (DeGCN).}
\textbf{Range} denotes the absolute difference between the highest and lowest accuracy for each action.}}
\label{tab:best_worst_actions}

\begin{minipage}[t]{0.497\textwidth}
\centering
\resizebox{\linewidth}{!}{
\begin{tabular}{lccccccr}
\toprule
\multicolumn{8}{c}{\textbf{Top-20 easiest actions (Ranked 1st - 20th)}} \\
\midrule
\textbf{Action} & \textbf{LA} & \textbf{STT} & \textbf{2s} & \textbf{ST} & \textbf{M-ST} & \textbf{De} & \textbf{Range} \\
\midrule
clapping each other & \textcolor{bestcolor}{98.4} & 98.4 & 92.6 & 95.5 & 95.2 & 96.4 & 5.8 \\
walk apart & 95.8 & 97.1 & 96.4 & 96.1 & \textcolor{bestcolor}{97.4} & 83.8 & 13.6 \\
bow & 92.0 & \textcolor{bestcolor}{95.1} & 88.0 & 92.6 & 91.4 & 88.3 & 7.1 \\
walk form apart to together & \textcolor{bestcolor}{95.0} & 93.4 & 93.4 & 90.3 & 90.3 & 90.6 & 4.7 \\
cheers and drink & 93.5 & \textcolor{bestcolor}{94.5} & 79.2 & 83.1 & 79.8 & 76.9 & 17.6 \\
rock-paper-scissors & \textcolor{bestcolor}{93.5} & 89.9 & 88.5 & 90.5 & 90.8 & 88.5 & 5.0 \\
nod head & 91.2 & 89.8 & 87.7 & \textcolor{bestcolor}{92.1} & 91.8 & 88.6 & 4.4 \\
arm circles & 86.4 & 86.1 & 87.4 & 84.8 & 77.0 & \textcolor{bestcolor}{90.6} & 13.6 \\
clapping and hushing & \textcolor{bestcolor}{90.2} & 88.7 & 83.0 & 86.9 & 85.1 & 84.2 & 7.2 \\
wave hand & \textcolor{bestcolor}{88.8} & 88.5 & 83.1 & 79.9 & 81.4 & 79.3 & 8.9 \\
knock over & 86.7 & \textcolor{bestcolor}{88.3} & 87.7 & 85.2 & 82.7 & 85.5 & 5.6 \\
exchange object & \textcolor{bestcolor}{87.4} & 82.8 & 72.0 & 73.5 & 77.4 & 78.0 & 15.4 \\
shake hands & 86.5 & 81.5 & \textcolor{bestcolor}{87.1} & 81.8 & 78.9 & 73.3 & 13.8 \\
pushing & \textcolor{bestcolor}{87.0} & 81.8 & 81.5 & 80.9 & 76.2 & 73.8 & 13.2 \\
arm swings & 84.5 & \textcolor{bestcolor}{86.1} & 78.6 & 80.9 & 85.4 & 76.4 & 9.7 \\
falling down & 83.9 & 80.3 & 82.2 & 78.3 & \textcolor{bestcolor}{85.5} & 80.9 & 7.2 \\
touch elbows & 81.1 & 81.4 & 70.2 & 79.4 & \textcolor{bestcolor}{84.7} & 77.0 & 14.5 \\
squat down & 81.6 & \textcolor{bestcolor}{84.2} & 75.0 & 77.7 & 75.3 & 71.1 & 13.1 \\
follow person & \textcolor{bestcolor}{84.0} & 75.6 & 78.6 & 68.6 & 76.6 & 70.6 & 15.4 \\
whisper & 70.2 & 73.5 & 74.4 & 81.2 & \textcolor{bestcolor}{83.2} & 70.9 & 13.0 \\
\bottomrule
\end{tabular}
}
\end{minipage}
\hfill
\begin{minipage}[t]{0.49\textwidth}
\centering
\resizebox{\linewidth}{!}{
\begin{tabular}{lccccccr}
\toprule
\multicolumn{8}{c}{\textbf{Bottom-20 most challenging actions (Ranked 83rd - 102nd)}} \\
\midrule
\textbf{Action} & \textbf{LA} & \textbf{STT} & \textbf{2s} & \textbf{ST} & \textbf{M-ST} & \textbf{De} & \textbf{Range} \\
\midrule
put object into bag & \textcolor{bestcolor}{45.6} & 33.0 & 13.6 & 31.1 & 20.4 & 31.1 & 32.0 \\
flick hair & 32.9 & \textcolor{bestcolor}{45.6} & 38.1 & 20.2 & 42.3 & 45.6 & 25.4 \\
hit with object & 36.6 & 42.3 & 42.3 & 39.0 & \textcolor{bestcolor}{44.1} & 39.0 & 7.5 \\
pinch body (not arm) & 38.0 & 34.7 & 34.3 & \textcolor{bestcolor}{43.2} & 34.0 & 27.1 & 16.1 \\
drink water & \textcolor{bestcolor}{43.2} & 38.4 & 34.5 & 38.4 & 32.1 & 34.8 & 11.1 \\
pinch face with two hands & 30.5 & 33.8 & 24.4 & 29.6 & \textcolor{bestcolor}{42.2} & 29.2 & 17.8 \\
stab person & 39.0 & 34.8 & \textcolor{bestcolor}{41.3} & 31.9 & 38.1 & 33.2 & 9.4 \\
thumb up & \textcolor{bestcolor}{40.7} & 28.3 & 36.0 & 35.0 & 25.9 & 25.3 & 15.4 \\
chop (cut) person & \textcolor{bestcolor}{40.1} & 40.1 & 21.2 & 29.3 & 32.3 & 37.1 & 18.9 \\
open bottle & 37.4 & 22.9 & 26.2 & 12.1 & 24.1 & \textcolor{bestcolor}{39.2} & 27.1 \\
make ok sign & \textcolor{bestcolor}{36.7} & 28.6 & 19.5 & 24.7 & 16.6 & 21.1 & 20.1 \\
yawn & 37.2 & 31.6 & 32.2 & \textcolor{bestcolor}{35.7} & 30.4 & 20.1 & 17.1 \\
make victory sign & 14.6 & 18.8 & 15.3 & 16.2 & 15.6 & \textcolor{bestcolor}{34.1} & 19.5 \\
self-cutting with knife & \textcolor{bestcolor}{33.2} & 30.2 & 31.7 & 21.5 & 32.3 & 27.8 & 11.7 \\
cutting paper & 20.4 & \textcolor{bestcolor}{33.1} & 17.5 & 17.8 & 19.4 & 22.6 & 15.6 \\
play magic cube & \textcolor{bestcolor}{32.4} & 22.8 & 14.7 & 10.6 & 13.5 & 12.8 & 21.8 \\
wipe face & 20.9 & 23.9 & 22.7 & 19.5 & \textcolor{bestcolor}{26.8} & 19.8 & 7.3 \\
take object out of bag & \textcolor{bestcolor}{25.6} & 23.3 & 18.8 & 11.7 & 21.0 & 14.6 & 13.9 \\
ball up paper & 23.3 & 14.0 & 13.7 & 7.8 & 21.4 & \textcolor{bestcolor}{24.2} & 16.4 \\
fold paper & 13.4 & 2.5 & 13.4 & 1.9 & 4.0 & \textcolor{bestcolor}{15.8} & 13.9 \\
\bottomrule
\end{tabular}
}
\end{minipage}

\end{table*}

\subsection{Action-Level Performance Trends on ANUBIS}

We conducted a comprehensive analysis of action-level performance patterns across all 102 action categories in ANUBIS using six models. 
To understand the performance spectrum, we examined the top-20 easiest and bottom-20 most challenging actions, revealing distinct patterns that illuminate the strengths and limitations of current skeleton-based approaches (See Table \ref{tab:best_worst_actions}). 

\textbf{What easy vs. hard classes reveal?}
Across models, the top-20 classes average 84.6\% accuracy, while the bottom-20 average only 27.5\%. 
High-accuracy actions (clapping each other 96-98\%, walk apart 96-97\%, bow up to 95.1\%) share clear, whole-body motion signatures and large spatial displacements that produce distinctive spatiotemporal patterns. In contrast, the hardest classes involve fine-motor, hand-centric, or human-object interactions (fold paper, ball up paper, play magic cube, open bottle), where skeletal streams alone underspecify semantics due to missing object context and limited finger fidelity. This split confirms that skeleton-only pipelines excel at macro body dynamics but underperform on micro-manipulation.

\textbf{Cross-model agreement as a stability signal.}
The Range column 
serves as a useful consensus proxy. It's smaller for easy classes 
and larger for hard classes, 
indicating difficult categories induce greater architectural sensitivity. Stable/easy classes include clapping each other (range 5.8\%), knock over (5.6\%), rock-paper-scissors (5.0\%), walk from apart to together (4.7\%), and nod head (4.4\%). 
By contrast, high-variance categories like put object into bag (32.0\%), open bottle (27.1\%), flick hair (25.4\%), and play magic cube (21.8\%) 
show strong model-specific behavior, suggesting that fusion design and local feature modeling, rather than modality count, drive the differences.

\textbf{Model-level 
robustness.} Averaged over the top-20, LA-GCN leads (87.9\%), followed by STTFormer (86.9\%), Motif-STGCN (84.3\%), ST-GCN (84.0\%), 2s-AGCN (83.3\%), and DeGCN (81.2\%).
On the bottom-20, LA-GCN again ranks first (32.1\%), then STTFormer (29.1\%), DeGCN (27.7\%), Motif-STGCN (26.8\%), 2s-AGCN (25.6\%), and ST-GCN (23.9\%). Notably, DeGCN has the smallest accuracy drop from top-20 to bottom-20 ($\sim$ 53.5 pp vs. 55-60 pp for others), indicating 
better resilience on difficult, fine-grained classes, even though its absolute accuracy remains lower on the easy set. This pattern hints that DeGCN’s part/decoupling bias helps when global motion cues are weak.

\textbf{Who wins where (per-class wins)?}
Counting ties as shared wins: in the top-20, LA-GCN leads with 8 wins (walk from apart to together, rock-paper-scissors, clapping and hushing), STTFormer has 5 (bow, knock over, squat down), and Motif-STGCN has 4 (walk apart, falling down, touch elbows, whisper).
In the bottom-20, LA-GCN again leads (8 wins), while DeGCN secures 4 wins (open bottle, make victory sign, ball up paper, fold paper), confirming its hand/object-centric strength. Results suggest LA-GCN offers best overall balance, while DeGCN excels on tail cases where local articulation dominates.

\textbf{Action taxonomy: dyadic vs. single-subject fine motor.}
Many top classes are dyadic or coordinated (clapping each other, exchange object, shake hands, pushing, follow person, whisper). The presence of a partner supplies relative pose and motion cues that are easier to encode in graphs/transformers, boosting separability. Conversely, bottom classes are mostly single-subject, object-centric, or subtle gestures (fold paper, ball up paper, play magic cube, make ok sign, yawn). Here, object state and finger articulation, both weakly represented in standard skeletons, are decisive. The outcome argues for object-aware and hand-aware augmentations, {e.g., explicit hand-pose subgraphs; 
object proxy nodes inferred from motion; or lightweight RGB cues fused selectively}.

\textbf{Where each architecture fails and why?}
Architectural biases surface in outliers. For instance, DeGCN underperforms on walk apart (83.8\%) relative to others ($\sim$96-97\%), suggesting some decoupled designs may under-leverage global displacement and long-range cross-person cues. STTFormer shows excellent performance on macro patterns (bow 95.1\%) but can collapse on fine manipulation (fold paper 2.5\%), hinting at insufficient high-precision local attention or challenges with ambiguous rear-view hand cues. Motif-STGCN, despite being older, excels on global, rhythmically coherent actions (walk apart 97.4\%, falling down 85.5\%, whisper 83.2\%), likely benefiting from recurring motion motifs. These contrasts suggest that ensembling or hybridizing (LA-GCN backbone + DeGCN-style local decoupling + motif priors) could yield gains.

\textbf{Practical guidance and future directions.}
For general deployment, LA-GCN is the most reliable performer across head-and-tail classes. For hand/object-heavy applications, consider mixing DeGCN-like local articulation modules or hand-focused subgraphs. The bottom classes make a strong case for: (i) Multi-modal enrichment: add lightweight object/context cues (RGB patches, object heatmaps) or learned object nodes linked to hands. (ii) Local attention: high-resolution, hand-centric attention (hierarchical GCNs/ Transformers, dilated temporal windows) to capture micro-gestures. (iii) Interaction modeling: explicit cross-person relational edges and contact events for dyadic actions. (iv) Curriculum/augmentation: viewpoint-hard negatives, hand-pose perturbations, and synthetic object-interaction variations to reduce overfitting to macro motion. Beyond Top-1/Top-5, track (i) per-class stability via Range, (ii) hand/object subset scores (tail classes), and (iii) dyadic vs. single-subject splits. Reporting these alongside overall accuracy makes progress on ANUBIS more diagnostic and reduces gains driven by already-easy, macro-motion categories. ANUBIS is a valuable resource for future research.

\subsection{Analysis of Feature Type Impact}

{We analyze feature-type effects on the top three models (see Appendix).} 

\textbf{Performance variation across actions.}
Examining the per-class accuracy distribution across our top-performing models reveals clear patterns: actions with large spatial displacement or distinctive global movement (walk apart, clapping each other, arm swings) 
achieve high accuracy, often exceeding 90\%. These generate strong spatiotemporal signatures easily captured by GCN and Transformer architectures.
Conversely, fine-grained actions with subtle local motions, especially hand/finger movements, achieve lower accuracy (fold paper, make victory sign, cutting paper). Discriminative cues concentrate in few distal joints, worsened by skeleton extraction limitations in finger tracking, causing misclassifications. Semantic ambiguity between similar skeletal patterns (different object manipulations) further reduces performance.
Dyadic interactive actions generally outperform single-subject fine motor tasks. Relative positioning, coordinated timing, and interaction cues provide additional discriminative information, reinforcing the need for multi-person relational modeling.

\textbf{Selective benefits of bone feature integration.}
Our evaluation shows Bone features produce consistent gains for only 7 of 102 actions across three models, highlighting task-selectivity where limb geometry and joint structure are critical. Sneeze and thumb up show average gains of +8.26\% and +7.86\%, where subtle arm-hand cues are decisive. High-mobility actions like running and follow person also see moderate gains, suggesting bone vectors encode gait patterns. However, improvement rarity ($\times$7\% of actions) indicates Bone integration requires selective, dynamic application based on action category.

\textbf{Mixed and model-dependent effects of motion features.}
Our analysis reveals mixed and model-dependent effects of motion features, with substantial boosts for certain architectures while harming others. Play magic cube gains +24.68\% for DeGCN and +12.82\% for STTFormer, yet LA-GCN drops -6.41\%. Similarly, stand up benefits STTFormer and DeGCN (both $>$+17\%) but slightly declines in LA-GCN (-2.39\%). This inconsistency stems from temporal modeling and fusion strategy differences, some architectures exploit motion magnitude effectively while others are disrupted by noise. Fine-motor actions like cutting paper show selective benefit, suggesting motion features better serve actions with distinct temporal rhythms but require careful integration.

\textbf{Actions harmed by additional features.}
Perhaps most surprisingly, our analysis identified actions where adding Bone or Motion features severely degrades performance, with Motion as primary culprit. Walk apart together and walk apart suffer average drops of -44.13\% and -35.81\%, with some models collapsing from $>$95\% to $<$35\% accuracy. Simple locomotion or static actions suffer when motion signals inject confounding noise, especially when speed variations, occlusion, or skeletal jitter mimic other patterns. Interactive actions (surrender, bite person) also degrade, implying motion cues mislead networks when core discriminative features are spatial configurations, not velocity patterns. Consistent harm across models indicates systematic vulnerability in current fusion pipelines where unfiltered motion signals overwhelm stable joint-based representations.

\textbf{Implications for feature fusion strategies.}
These findings reveal a complex pattern: Bone features are occasionally helpful but mostly neutral, while Motion features are high-risk, high-reward, delivering substantial gains or catastrophic losses. This suggests adaptive, category-aware fusion strategies that adjust feature contributions based on action type, scene context, or model confidence.
Motion integration could be prioritized for dynamic temporal actions (jump up, play magic cube) while disabled for stable-pose actions (walk apart). Bone features could target fine-hand gestures or limb-orientation-dependent actions. Future architectures could use attention-based feature weighting or meta-learning frameworks to dynamically select which features to emphasize per instance.
{
These trends are consistent with the geometric interpretation. Joint features are most reliable when action recognition depends primarily on global body configuration or coarse spatiotemporal layout. Bone features are useful when relative limb orientation and local structural constraints are discriminative, but their gains are naturally selective because they overlap with part of the information already encoded by joints. Motion features behave differently: since they are temporal increments rather than state descriptors, they are most helpful when the action is characterized by distinctive phase transitions or motion rhythms, but can become harmful when the action is dominated by stable poses, slow displacement, or noisy temporal variation. This explains why adding motion may improve some fine-grained dynamic classes while degrading others, and why simply combining more feature types does not guarantee better recognition. In essence, the success of fusion depends not on the number of representations, but on whether their geometric properties, noise characteristics, and semantic emphasis are aligned with the target action.
}

\subsection{{Future Research Directions}}

{Our in-depth evaluation of ANUBIS highlights persistent limitations in current skeleton-based action recognition approaches, particularly in how they handle heterogeneous action types, feature integration, and real-world deployment. These limitations show opportunities for fundamental advances. }

{\textbf{Adaptive feature-type fusion.} Our results show that the utility of different feature types, joint coordinates, bone vectors, and motion cues, varies dramatically between action categories. For example, large-scale displacement actions like walk apart achieve near-saturation accuracy with joint alone, whereas fine-grained manipulations such as cutting paper depend far more on high-frequency motion cues. However, most existing fusion pipelines treat all features as equally relevant, which can dilute useful signals and even reduce performance.}

{Future work should replace static fusion with dynamic feature-type selection. Models could learn action-feature relevance mappings via attention gates or controller networks, activating only the most informative features per action instance or even per temporal phase (e.g., preparation, execution, recovery). Beyond selection, fusion should respect semantic hierarchy: joint often encodes high-level pose, bone captures execution strategies, and motion reflects low-level motion dynamics. Explicitly modeling this hierarchy, potentially with language-aligned semantic anchors, could enable richer cross-feature reasoning and improve both interpretability and generalization.}

{\textbf{Large language model-driven action understanding.} The structured nature of skeleton data makes it ideal for multimodal large language models (MLLMs). While models like CLIP excel in image-text alignment, their temporal reasoning remains weak. ANUBIS, rich in multi-view, socially interactive, and modern behavior patterns, offers fertile ground for skeleton-informed temporal encoders that strengthen video-language alignment.
Key directions include skeleton-text contrastive learning for fine-grained action semantics, skeleton-guided attention mechanisms within MLLMs, and tri-modal fusion of skeleton, video, and text. Skeletons bring unique advantages: robustness to background/lighting, compact representation for long sequences, and built-in privacy preservation. These traits make them valuable for surveillance, instructional content generation, and accessibility tools. Notably, ANUBIS's rear-view and complex dyadic interactions could help train models that understand subtle interpersonal cues often missed in RGB-based training.}

{\textbf{Social safety and behavioral dynamics analysis.} ANUBIS’s diverse interpersonal scenarios create new opportunities for safety-critical applications like bullying or harassment detection. Skeleton-based systems can quantify spatial dynamics, role asymmetries, and temporal escalation patterns without exposing identifiable visual information. This enables early detection of harmful behaviors while preserving privacy.}
{Further, the dataset’s coverage of pandemic-era social norms (e.g., elbow touches, distancing gestures) enables studying cultural and temporal shifts in interaction patterns. Combining skeletal kinematics with interaction graphs could yield models capable of detecting subtle power imbalances, changes in group cohesion, or shifts in emotional state, all valuable for workplace monitoring, school safety, and public health.}

{\textbf{Healthcare, rehabilitation, and cognitive monitoring.}
Skeleton-based recognition offers an objective, non-invasive framework for medical and wellness applications. In rehabilitation, precise joint tracking enables quantitative movement quality assessment, replacing subjective clinician scoring. Tailored systems could detect Parkinson’s tremors, track stroke recovery symmetry, or adapt training regimens dynamically.}
{In eldercare, gait and balance analysis can feed predictive models for fall prevention, while continuous monitoring of daily activity patterns supports independent living assessments. Beyond physical health, subtle deviations in movement rhythm, coordination, or social behavior may serve as early markers for cognitive decline or mental health issues. The social interaction data in ANUBIS could be leveraged to build behavioral baselines for early intervention in schools, workplaces, and care facilities.}

\section{Conclusion}

{This work presents a representation-centric review of skeleton-based action recognition, introduces the challenging 
benchmark, and systematically evaluates state-of-the-art models across joint, bone, and motion features. Our results show that multi-representational fusion is not universally beneficial: bone features effectively capture fine-grained geometric coordination, while motion features are advantageous for cyclic or phase-transition actions, yet can significantly degrade performance for large-scale displacements, synchronized interactions, and static poses. We attribute this to the heterogeneous geometries of the feature spaces, including Euclidean space for joints, Lie group structure for bones, and tangent space for motion, where na\"ive fusion introduces redundancy, noise, and semantic inconsistency.
These findings challenge the assumption that adding more representations inherently improves recognition. Instead, effective action recognition requires task-aware, semantically aligned fusion strategies that respect feature compatibility. Beyond this analysis, ANUBIS contributes a diverse 
benchmark featuring back-view actions, violent interactions, and complex multi-person dynamics, exposing current model limitations while providing a foundation for developing more adaptive, generalizable, and context-aware recognition systems.}

\appendix

\definecolor{bestcolor}{HTML}{BE830E}
\definecolor{ANUGold}{HTML}{BE830E}
\definecolor{ANUGoldTint}{HTML}{F5EDDE}
\definecolor{ANUGoldDark}{HTML}{9A6B0B}
\definecolor{ANUGoldLight}{HTML}{D4A656} 
\definecolor{ANUUnigrey}{HTML}{333333}
\definecolor{ANUBlack}{HTML}{000000}  
\definecolor{best}{HTML}{ED6F6E} 
\definecolor{second}{HTML}{5560AC} 




\section{Comparison with Existing Datasets}
Table \ref{tab:skeleton_datasets} presents a comparative summary of our dataset against existing benchmarks.
{Existing datasets exhibit diverse characteristics across devices, modalities, and applications. NTU-60~\cite{2016_cvpr_ntu} establishes a multi-view benchmark with 60 indoor actions, while NTU-120~\cite{2019_tpami_ntu120} extends to 120 classes. Kinetics-skeleton~\cite{2018_aaai_st_gcn} extracts skeletal information from large-scale RGB videos. For multimodal fusion, PKU-MMD I~\cite{PKU-MMD-I} integrates RGB, depth, and infrared data, while PKU-MMD II~\cite{PKU-MMD-II} adds fine-grained interaction annotations. MMAct~\cite{mmact} combines visual and inertial sensor data for mobile scenarios, and RGB-D Varying-View~\cite{RGB-D-Varying-View} explores viewpoint robustness through dynamic perspectives.In vertical applications, ETRI-Activity 3D~\cite{ETRI-Activity3D} focuses on elderly monitoring, IKEA ASM~\cite{IKEA-ASM} targets furniture assembly tasks, NCRC-Human~\cite{NCRC} analyzes nursing scenarios, and Tai-Chi~\cite{Tai-Chi} characterizes Tai Chi kinematics. UAV-Human~\cite{UAV-Human} enriches spatial dimensions through drone perspectives. These datasets serve different objectives including general validation, scenario-specific optimization, and cross-modal learning, forming a diversified research ecosystem.}

\begin{table*}[htbp]
\setlength{\tabcolsep}{0.15em}
\renewcommand{\arraystretch}{0.70}
\centering
\caption{Comprehensive Overview of Skeleton-based Action Recognition Datasets. ANUBIS addresses critical gaps through rear-view coverage, aggressive/security actions, and enhanced 32-joint representation with Azure Kinect.}
\label{tab:skeleton_datasets}
\resizebox{\textwidth}{!}{%
\begin{tabular}{lccclllll}
\toprule
\textbf{Dataset} & \textbf{Classes} & \textbf{Views} & \textbf{Subjects} & \textbf{Clips} & \textbf{Sensors} & \textbf{Additional Modalities} & \textbf{Dataset Type} \\
\midrule
HDM05 (2007)\cite{hdm05} & 130 & 1  & 5 & 2,337 & - & RGB & Human motion capture \\
MSRAction3D (2010)\cite{MSRAction3D} & 20 & 1 & 10 & 567 & Kinect & RGB, Depth & Daily activities \\
CAD-60 (2011)\cite{cad-60} & 12 & - & 4 & 68 & Kinect & RGB, Depth & Human performing activities \\
MSRDailyActivity3D (2012)\cite{MSRDailyActivity3D} & 16 & 1 & 10 & 320 & Kinect & RGB, Depth & Daily activities \\
G3D-Gaming (2012)\cite{G3D-Gaming} & 20 & 1 & 10 & - & Kinect & RGB, Depth & Gaming gestures \\
UTKinect (2012)\cite{UTKinect-Action3D} & 10 & - & 10 & 200 & Kinect & RGB, Depth & Human actions \\
SBU (2012)\cite{sbu} & 8 & - & 7 & 282 & Kinect & RGB & Human-human interaction \\
CAD-120 (2013)\cite{cad120} & 10 & - & 4 & 120 & Kinect & RGB, Depth & Activity types \& object interactions \\
Berkeley MHAD (2013)\cite{Berkeley-MHAD} & 11 & 4 & 12 & 660 & Kinect & RGB, Depth, Audio, Accelerometer & Multimodal Capture \& Controllable \& Synced Data \\
Florence3D-Action (2013)\cite{Florence3D-Action} & 9 & - & 10 & 215 & Kinect & RGB, Depth & Daily Activities \\
MSRActionPairs3D (2013)\cite{MSRActionPairs3D} & 12 & - & 10 & 360 & Kinect & RGB, Depth & 3D Action \& Gesture Recognition \\
UCFKinect (2013)\cite{UCFKinect} & 16 & - & 16 & 1,280 & - & RGB, Depth & General actions \\
Northwestern-UCLA (2014)\cite{Northwestern-UCLA} & 10 & 3 & 10 & 1,494 & Kinect & RGB, Depth & Daily Activities \\
Multi-View TJU (2014)\cite{Multi-View-TJU} & 20 & 2 & 22 & 7,040 & - & RGB, Depth & Multi-view actions \\
UWA3D Multiview Activity (2014)\cite{UWA3D-Multiview-Activity} & 30 & 4 & 10 & 701 & Kinect & RGB, Depth & Multi-view actions \\
SYSU 3D HOI (2015)\cite{sysu} & 12 & - & 40 & 480 & Kinect & RGB, Depth & Human-object interaction \\
UWA3D Multiview Activity II (2015)\cite{UWA3D-Multiview-Activity-II} & 30 & 4 & 10 & 1,070 & Kinect & RGB, Depth & Daily activities \\
NTU-60 (2016)\cite{2016_cvpr_ntu} & 60 & 80 & 40 & 56,880 & Kinect v2 & RGB, Depth, Infrared & Large-scale general actions \\
PKU-MMD I (2017)\cite{PKU-MMD-I} & 51 & 3 & 66 & 1,076 & Kinect v2 & RGB, Depth, Infrared & Multi-modal actions \\
Kinetics-skeleton (2018)\cite{2018_aaai_st_gcn} & 400 & - & - & 260,232 & - & - & Based on publicly available RGB videos \\
RGB-D Varying-View (2018)\cite{RGB-D-Varying-View} & 40 & 9 & 118 & 25,600 & Kinect v2 & RGB, Depth & Multi-view actions \\
NTU-120 (2019)\cite{2019_tpami_ntu120} & 120 & 155 & 106 & 114,480 & Kinect v2 & RGB, Depth, Infrared & Large-scale general actions \\
MMAct (2019)\cite{mmact} & 37 & 5 & 20 & 36,764 & - & RGB, Accelerometer, Gyroscope & Multi-modal actions \\
PKU-MMD II (2020)\cite{PKU-MMD-II} & 41 & 3 & 13 & 1,009 & Kinect v2 & RGB, Depth, Infrared & Multi-modal actions \\
ETRI-Activity3D (2020)\cite{ETRI-Activity3D} & 55 & - & 100 & 112,620 & Kinect v2 & RGB & Daily activities of the elderly \\
IKEA ASM (2020)\cite{IKEA-ASM} & 33 & 3 & 48 & 16,764 & Kinect v2 & RGB, Depth & Furniture assembly \\
UAV-Human (2021)\cite{UAV-Human} & 155 & - & 119 & 22,476 & Azure Kinect & RGB, Infrared, Depth& UAV perspective actions \\
NCRC (2022)\cite{NCRC} & 6 & - & 8 & 398 & - & - & Nursing care activities \\
Tai-Chi (2022)\cite{Tai-Chi} & 10 & - & - & 200 & Perception Neuron & - & Martial arts \\
ANUBIS (2025) & 102 & 80 & 80 & 66,232 & Azure Kinect & RGB, Depth & Large-Scale \& Multi-Person \& Frontal / Rear-View \& In-the-Wild \\
\bottomrule
\end{tabular}
}
\end{table*}


\section{Dataset Collection}

\begin{figure}[htbp]
    \centering
    \begin{minipage}[b]{0.495\linewidth}
        \centering
        \includegraphics[width=0.86\linewidth]{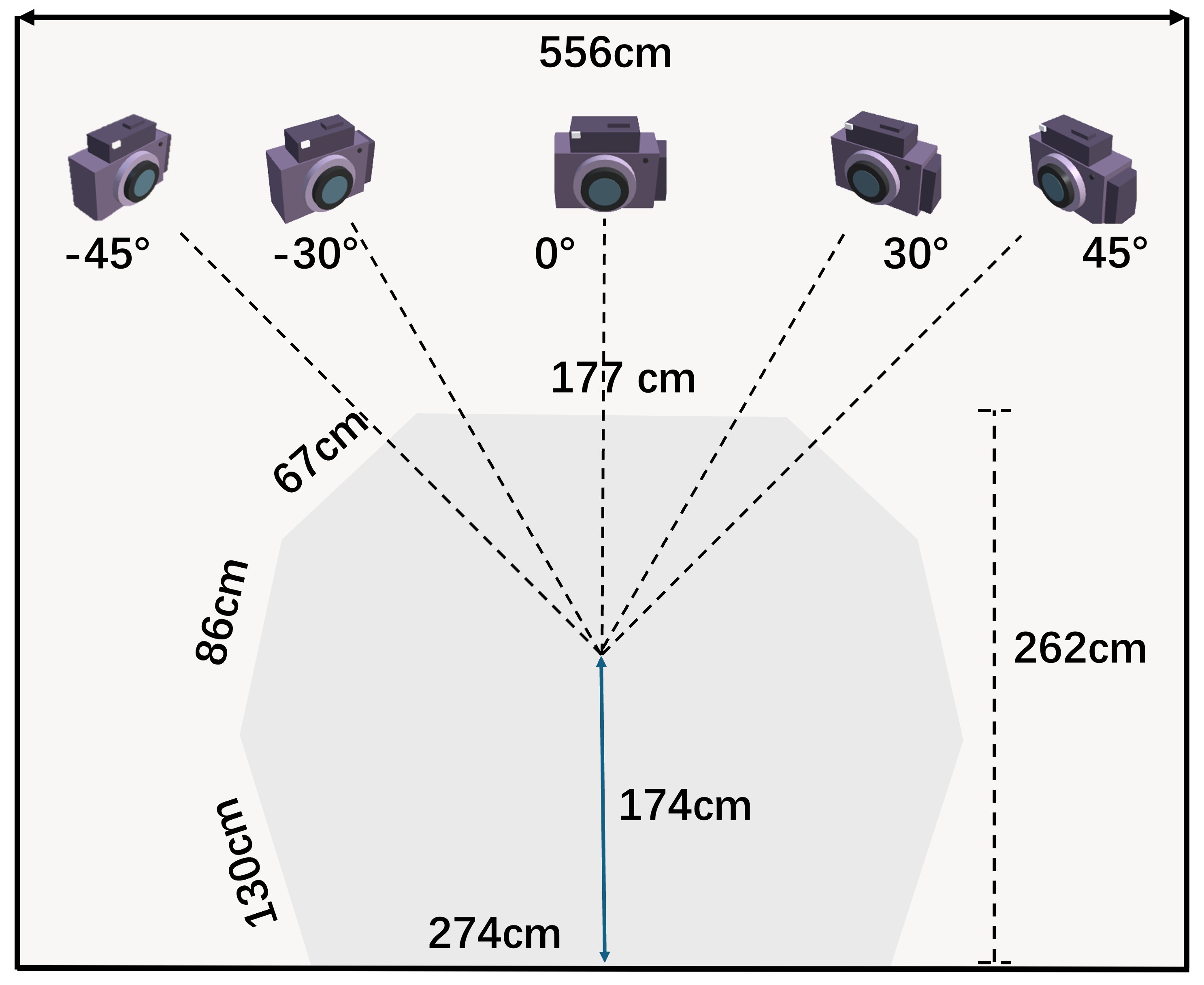}
        \caption*{(a) Venue layout.} 
        \label{fig:minipage1}
    \end{minipage}\hfill
    \begin{minipage}[b]{0.495\linewidth}
        \centering
        \includegraphics[width=0.975\linewidth]{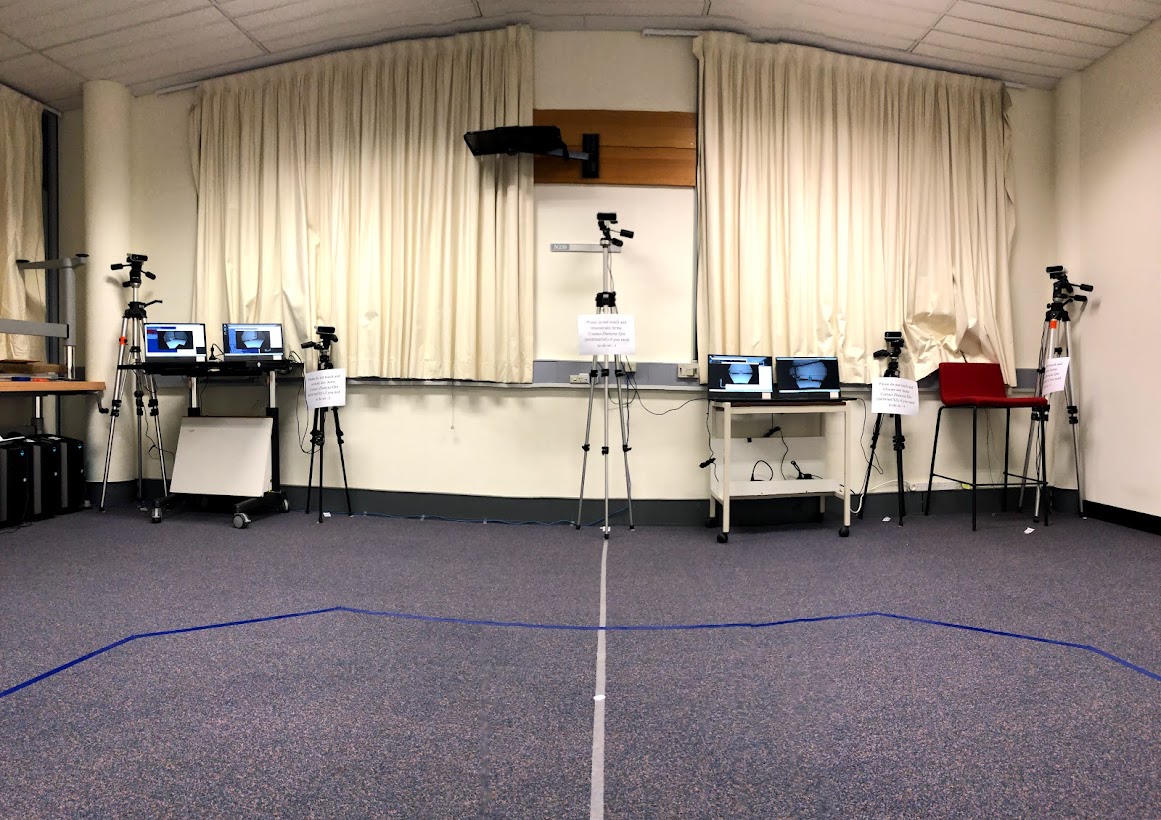}
        \caption*{(b) Camera arrangement.} 
        \label{fig:minipage2}
    \end{minipage}
    \caption{ANUBIS dataset collection setup overview.}
    \label{fig:combined}
\end{figure}

ANUBIS comprises 102 carefully selected actions including both individual behaviors (e.g., drinking water, waving) and multi-person interactions (e.g., handshaking, object exchange, stabbing). The 102 actions are distributed across 40 collection sessions, with each session involving two participants as a group and lasting approximately 1.5 hours. Every 10 sessions incorporates 10-minute breaks to maintain performance quality, and sessions exhibiting substandard action execution are re-recorded to ensure data integrity. The dataset comprises 40 participant groups totaling 80 participants and approximately 60 hours of multi-modal recordings
(see Fig. 1 in the main paper). 

\textbf{Multi-view camera setups.} The acquisition system uses five Microsoft Azure Kinect devices arranged in symmetric horizontal configuration at $0^\circ$, $\pm30^\circ$, and $\pm45^\circ$ angles within a standardized 556cm$\times$274cm indoor environment, as illustrated in Fig.~\ref{fig:combined}. While cameras maintain horizontal symmetry, each device operates at different heights and poses to enhance viewpoint diversity. Camera heights and poses are randomly adjusted every 10 groups to capture more diverse viewpoints.

Participants move freely within marked activity areas and perform each action four times per session to create different viewpoints: facing the cameras, facing away from cameras, switching positions while facing cameras, and switching positions while facing away. This collection protocol results in 20 different camera views for each participant pair performing the same action, ensuring comprehensive coverage from multiple angles, especially challenging rear views that are often missing in existing datasets. For interactive actions, participants also switch their active and passive roles when changing positions to capture both sides of the interaction.
To ensure realistic performances while maintaining safety, we use appropriate items for different action types: toy weapons for simulated violence, wigs for hair-pulling actions, tissues for mouth-covering gestures, and soft objects like paper boxes for hitting actions to prevent injury.

\textbf{Data preprocessing.}
We developed custom software to manage data collection across all five synchronized Azure Kinect cameras. The software records the exact start and end time of each action, ensuring all cameras capture the same actions simultaneously.
During data processing, we use these recorded timestamps to extract individual action clips from the complete recordings of each group. Each clip contains three types of data: RGB, depth, and 3D skeleton videos, as shown in Fig. 1 in the main paper.
For actions that are naturally short, we extend them to the standard 300-frame length by repeating the action frames.                         

\begin{figure}[htbp]
    \centering
    \includegraphics[width=\linewidth]{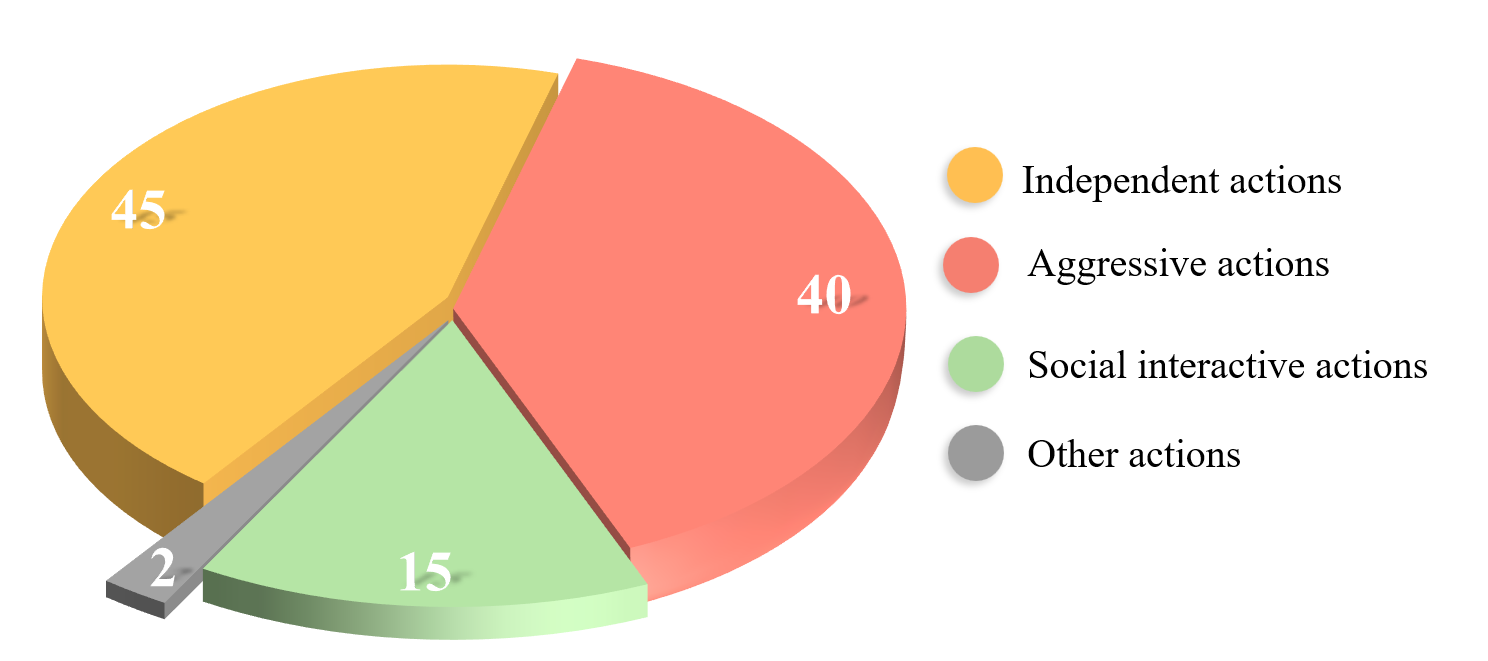}
    \caption{Distribution of 102 human actions classified into four categories. The pie chart shows: independent actions (45, 44.1\%), aggressive actions (40, 39.2\%), social interactive actions (15, 14.7\%), and other actions (2, 2.0\%). Other actions specifically refer to spatial position change behaviors, including walk apart and walk from apart to together.
}
    \label{fig:piechart}
\end{figure}

\textbf{Dataset statistics.} ANUBIS comprises 102 action categories collected from 80 participants, generating 66,232 skeleton clips across 80 viewpoints, as presented in Tab.\ref{tab:skeleton_datasets}. The viewpoint distribution includes 40 frontal views and 40 rear views from different angles, ensuring balanced coverage between frontal perspectives and challenging posterior orientations. Based on action categories, the dataset contains 45 independent actions (single-person behaviors) and 57 multi-person interactions. Among multi-person actions, we include 17 social interaction behaviors (e.g., handshaking, patting shoulders, object exchange), and 40 aggressive actions (e.g., hitting, stabbing, strangling). The complete statistics of ANUBIS are shown in Fig.\ref{fig:piechart}.

\section{Experimental Setups}

We benchmarked a range of state-of-the-art skeleton-based action recognition methods on our newly collected ANUBIS dataset and evaluated these methods on the NTU datasets for comparative analysis.
All experiments were implemented in PyTorch and trained on a single NVIDIA RTX 3090 GPU for 50 epochs. 
Stochastic Gradient Descent (SGD) with momentum 0.9 was used as the optimizer, with an initial learning rate of 0.05, decayed to 10\% at epoch 30.  

Skeleton data was preprocessed via normalization and translation. All video clips were standardized to 300 frames using action repetition (except SkateFormer~\cite{2024_ECCV_Skateformer}, which retained its original 64-frame window). 
%
Evaluation metrics included Top-1 and Top-5 classification accuracy, as well as model complexity indicators. To show the recognition accuracies of a model for all the action classes, a confusion matrix is used \cite{lei_tip_2019}.

\section{Additional Analysis and Discussion}

\textbf{Per-class performance analysis: easy vs. hard actions.} This section presents a detailed breakdown of action recognition performance across the 102 action categories in ANUBIS, revealing clear patterns in what makes certain actions easy or challenging for current skeleton-based models. 

\textbf{Bone feature integration analysis.} This section examines which specific actions benefit from adding bone vector features to joint coordinates, revealing the selective nature of anatomical structure information. The detailed results are presented in Table \ref{tab:bone_improvements}.

\begin{table*}[tbp]
\centering
\caption{Actions with consistent accuracy gains across all three models (LA-GCN, STTFormer, DeGCN) when adding the Bone feature type. Only 7 out of 102 classes show universal benefit, indicating that bone vectors selectively help actions where limb geometry and joint relationships are key (e.g., arm-hand positioning).}
\label{tab:bone_improvements}
\resizebox{\textwidth}{!}{%
\begin{tabular}{@{}clccccccccccc@{}}
\toprule
\multirow{2}{*}{\textbf{Rank}} & \multirow{2}{*}{\textbf{Action Name}} & \multicolumn{3}{c}{\textbf{LA-GCN}} & \multicolumn{3}{c}{\textbf{STTFormer}} & \multicolumn{3}{c}{\textbf{DeGCN}} & \multirow{2}{*}{\textbf{Avg. Improve}} \\
\cmidrule(lr){3-5} \cmidrule(lr){6-8} \cmidrule(lr){9-11}
& & \textbf{Joint} & \textbf{Joint+Bone} & \textbf{Improve} & \textbf{Joint} & \textbf{Joint+Bone} & \textbf{Improve} & \textbf{Joint} & \textbf{Joint+Bone} & \textbf{Improve} & \\
\midrule
1 & sneeze & 0.5310 & 0.5693 & +0.0383 & 0.3628 & 0.5192 & +0.1564 & 0.4779 & 0.5310 & +0.0531 & \textbf{+0.0826} \\
2 & thumb up & 0.4074 & 0.4175 & +0.0101 & 0.2828 & 0.4141 & +0.1313 & 0.2525 & 0.3468 & +0.0943 & \textbf{+0.0786} \\
3 & follow person & 0.8395 & 0.8696 & +0.0301 & 0.7559 & 0.8462 & +0.0903 & 0.7057 & 0.7559 & +0.0502 & \textbf{+0.0569} \\
4 & running & 0.7883 & 0.7948 & +0.0065 & 0.7362 & 0.8241 & +0.0879 & 0.7557 & 0.8013 & +0.0456 & \textbf{+0.0467} \\
5 & pull collar & 0.4202 & 0.4448 & +0.0246 & 0.3834 & 0.4141 & +0.0307 & 0.3006 & 0.3804 & +0.0798 & \textbf{+0.0450} \\
6 & self-cutting with knife & 0.3323 & 0.3867 & +0.0544 & 0.3021 & 0.3263 & +0.0242 & 0.2779 & 0.2870 & +0.0091 & \textbf{+0.0292} \\
7 & punch to face & 0.4787 & 0.5488 & +0.0701 & 0.4909 & 0.4939 & +0.0030 & 0.3994 & 0.4116 & +0.0122 & \textbf{+0.0284} \\
\bottomrule
\end{tabular}%
}
\end{table*}

\textbf{Motion feature integration analysis.} This section analyzes the impact of adding motion features (velocity/acceleration) to skeleton representations, revealing highly variable and model-dependent effects. Table \ref{tab:motion_improvements} presents the detailed analysis.

\begin{table*}[tbp]
\centering
\caption{Actions improved in at least two of three models when adding the Motion feature type. No action achieved consistent gains across all models, underscoring the model-dependent and unstable nature of motion integration, beneficial for certain temporally distinctive actions but harmful in others. Action abbreviations: ``support old people walking'' refers to ``support with arms for old people walking''.}
\label{tab:motion_improvements}
\resizebox{\textwidth}{!}{%
\begin{tabular}{@{}clccccccccccc@{}}
\toprule
\multirow{2}{*}{\textbf{Rank}} & \multirow{2}{*}{\textbf{Action Name}} & \multicolumn{3}{c}{\textbf{LA-GCN}} & \multicolumn{3}{c}{\textbf{STTFormer}} & \multicolumn{3}{c}{\textbf{DeGCN}} & \multirow{2}{*}{\textbf{Avg. Improve}} \\
\cmidrule(lr){3-5} \cmidrule(lr){6-8} \cmidrule(lr){9-11}
& & \textbf{Joint} & \textbf{Joint+Motion} & \textbf{Improve} & \textbf{Joint} & \textbf{Joint+Motion} & \textbf{Improve} & \textbf{Joint} & \textbf{Joint+Motion} & \textbf{Improve} & \\
\midrule
1 & stand up & 0.7964 & 0.7725 & -0.0239 & 0.4820 & 0.6617 & +0.1797 & 0.5719 & 0.7605 & +0.1886 & \textbf{+0.1148} \\
2 & play magic cube & 0.3237 & 0.2596 & -0.0641 & 0.2276 & 0.3558 & +0.1282 & 0.1282 & 0.3750 & +0.2468 & \textbf{+0.1036} \\
3 & take off a hat & 0.5761 & 0.7104 & +0.1343 & 0.5493 & 0.5284 & -0.0209 & 0.3851 & 0.4567 & +0.0716 & \textbf{+0.0617} \\
4 & play a phone & 0.4149 & 0.5403 & +0.1254 & 0.4239 & 0.3403 & -0.0836 & 0.2657 & 0.3791 & +0.1134 & \textbf{+0.0517} \\
5 & cutting paper & 0.2038 & 0.3726 & +0.1688 & 0.3312 & 0.3121 & -0.0191 & 0.2261 & 0.2261 & +0.0000 & \textbf{+0.0499} \\
6 & running & 0.7883 & 0.8241 & +0.0358 & 0.7362 & 0.8143 & +0.0781 & 0.7557 & 0.7362 & -0.0195 & \textbf{+0.0315} \\
7 & support old people walking & 0.7800 & 0.6967 & -0.0833 & 0.6500 & 0.7067 & +0.0567 & 0.5700 & 0.6700 & +0.1000 & \textbf{+0.0245} \\
8 & squat down & 0.8155 & 0.8452 & +0.0297 & 0.8423 & 0.7887 & -0.0536 & 0.7113 & 0.8065 & +0.0952 & \textbf{+0.0238} \\
9 & jump up & 0.7545 & 0.7455 & -0.0090 & 0.6108 & 0.7156 & +0.1048 & 0.7126 & 0.6826 & -0.0300 & \textbf{+0.0219} \\
10 & pull collar & 0.4202 & 0.3957 & -0.0245 & 0.3834 & 0.4325 & +0.0491 & 0.3006 & 0.3712 & +0.0706 & \textbf{+0.0317} \\
\bottomrule
\end{tabular}%
}
\end{table*}

\textbf{Negative impact analysis: when additional features hurt performance.} This section identifies actions where adding bone or motion features consistently degrades performance across all models, highlighting potential pitfalls in naive feature fusion. The comprehensive results are shown in Table \ref{tab:negative_impacts}.

\begin{table*}[tbp]
\centering
\caption{Actions with consistent accuracy drops across all three models when adding either Bone or Motion features (worst-affected feature type reported). All top declines are linked to Motion, with some drops exceeding 40\%, highlighting the risk of unfiltered motion cues overwhelming stable joint-based representations. Action abbreviations: ``walk apart together'' refers to ``walk form apart to together'' and ``throw object to person'' refers to ``pick and throw an object to person''.}
\label{tab:negative_impacts}
\resizebox{\textwidth}{!}{%
\begin{tabular}{@{}clcccccccccccc@{}}
\toprule
\multirow{2}{*}{\textbf{Rank}} & \multirow{2}{*}{\textbf{Action Name}} & \multirow{2}{*}{\textbf{Feature}} & \multicolumn{3}{c}{\textbf{LA-GCN}} & \multicolumn{3}{c}{\textbf{STTFormer}} & \multicolumn{3}{c}{\textbf{DeGCN}} & \multirow{2}{*}{\textbf{Avg. Decline}} \\
\cmidrule(lr){4-6} \cmidrule(lr){7-9} \cmidrule(lr){10-12}
& & & \textbf{Joint} & \textbf{Added feature} & \textbf{Decline} & \textbf{Joint} & \textbf{Added feature} & \textbf{Decline} & \textbf{Joint} & \textbf{Added feature} & \textbf{Decline} & \\
\midrule
1 & walk apart together & Motion & 0.9497 & 0.8365 & -0.1132 & 0.9340 & 0.2799 & -0.6541 & 0.9057 & 0.3491 & -0.5566 & \textbf{-0.4413} \\
2 & walk apart & Motion & 0.9579 & 0.3042 & -0.6537 & 0.9709 & 0.8123 & -0.1586 & 0.8382 & 0.5761 & -0.2621 & \textbf{-0.3581} \\
3 & surrender & Motion & 0.7212 & 0.6154 & -0.1058 & 0.7308 & 0.4872 & -0.2436 & 0.7372 & 0.4423 & -0.2949 & \textbf{-0.2148} \\
4 & bite person & Motion & 0.6592 & 0.5732 & -0.0860 & 0.6369 & 0.3949 & -0.2420 & 0.6561 & 0.3631 & -0.2930 & \textbf{-0.2070} \\
5 & fist bumping & Motion & 0.8190 & 0.6499 & -0.1691 & 0.7774 & 0.5905 & -0.1869 & 0.5816 & 0.3917 & -0.1899 & \textbf{-0.1820} \\
6 & back pain & Motion & 0.5131 & 0.4739 & -0.0392 & 0.6176 & 0.4216 & -0.1960 & 0.5817 & 0.2745 & -0.3072 & \textbf{-0.1808} \\
7 & throw object to person & Motion & 0.6730 & 0.5143 & -0.1587 & 0.7143 & 0.5143 & -0.2000 & 0.5556 & 0.3778 & -0.1778 & \textbf{-0.1788} \\
8 & open bottle & Motion & 0.3735 & 0.2018 & -0.1717 & 0.2289 & 0.1506 & -0.0783 & 0.3916 & 0.1325 & -0.2591 & \textbf{-0.1697} \\
9 & thumb down & Motion & 0.5623 & 0.4815 & -0.0808 & 0.6364 & 0.3939 & -0.2425 & 0.5320 & 0.3906 & -0.1414 & \textbf{-0.1549} \\
10 & strangling neck & Motion & 0.5666 & 0.3746 & -0.1920 & 0.4799 & 0.3715 & -0.1084 & 0.4458 & 0.2817 & -0.1641 & \textbf{-0.1548} \\
\bottomrule
\end{tabular}%
}
\end{table*}

\begin{figure*}[htbp]
    \centering
    \includegraphics[width=\textwidth]{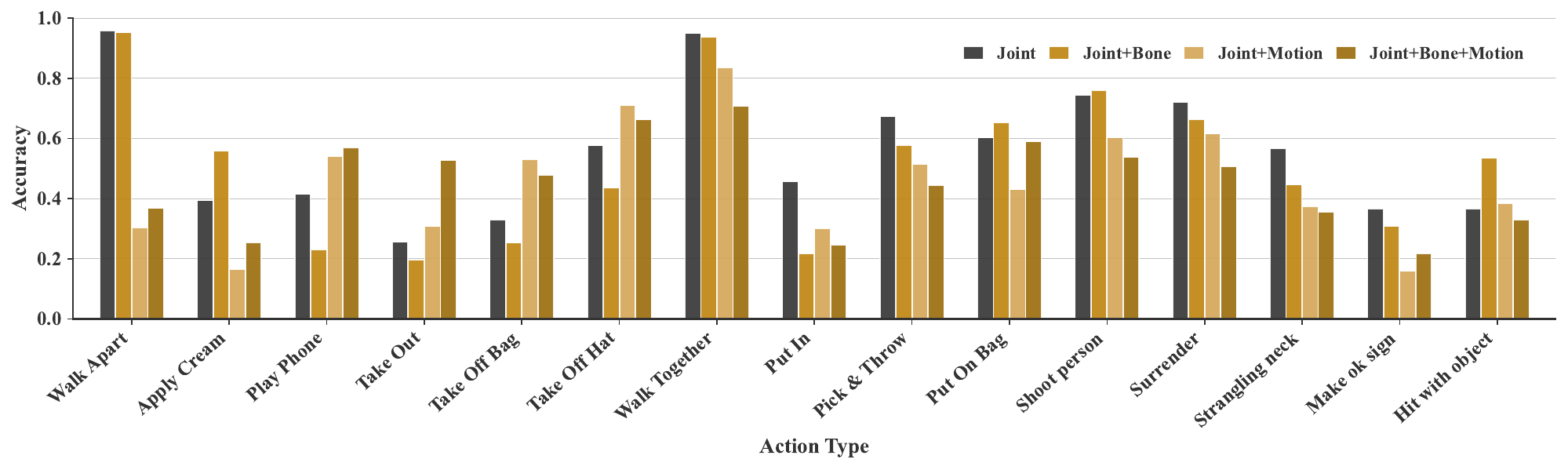}
    \caption{Analysis of Joint, Bone, Motion data stream effects on action recognition using LA-GCN. 
Results from 15 actions with significant recognition accuracy fluctuations are shown. 
Action abbreviations: Apply Cream (apply cream on hand), Take Out (take object out of bag), 
Walk Together (walk form apart to together), Put In (put object into bag), 
Pick \& Throw (pick and throw an object to person), Support Walk (support with arms for old people walking).}
    \label{fig:accuracy_comparison_15}
\end{figure*}

\textbf{Feature stream effect visualization.} This section provides a visual analysis of how different feature combinations affect action recognition performance, illustrating the complex interplay between joint, bone, and motion representations. The results are visualized in Figure \ref{fig:accuracy_comparison_15}.

The visualization reveals several key patterns:

\renewcommand{\labelenumi}{\roman{enumi}.}
\begin{enumerate}
    \item {Feature interaction effects}: Some actions benefit from bone features but are harmed by motion (e.g., Walk Apart), while others show the opposite pattern (e.g., Apply Cream).
    \item {Non-additive fusion}: The best performance often comes from selective feature combinations rather than using all available features. For instance, "Walk Together" performs best with Joint+Bone but degrades significantly when motion is added.
    \item {Action-specific optimization}: Different actions require different feature strategies, suggesting the need for adaptive or action-aware fusion mechanisms rather than universal multi-modal approaches.
    \item {Complementary vs. competing features}: While some feature combinations are complementary (Joint+Bone for spatial actions), others compete or introduce noise (Motion for stable pose actions like Surrender).
\end{enumerate}

\small{
\bibliographystyle{plain}
\bibliography{reference.bib}}


\end{document}